\documentclass{article}

\usepackage{epsfig}
\usepackage{amssymb}
\usepackage{graphicx}
\usepackage{subfig}
\usepackage[round]{natbib}

\usepackage{hyperref}       % hyperlinks
\usepackage{url}            % simple URL typesetting
\usepackage{booktabs}       % professional-quality tables
\usepackage{amsfonts}       % blackboard math symbols
\usepackage{nicefrac}       % compact symbols for 1/2, etc.
\usepackage{microtype}      % microtypography
\usepackage{xcolor}         % colors

\usepackage{algorithm}
\usepackage{algorithmic}
\usepackage{amsmath}
\usepackage{mathtools}
\usepackage[capitalize,noabbrev]{cleveref}

\usepackage[textsize=tiny]{todonotes}
\usepackage[utf8]{inputenc}
\usepackage[acronym,nowarn]{glossaries}
\glsdisablehyper
\makeglossaries
\usepackage{color}
\usepackage{cancel}

\usepackage[accepted]{icmldraft}
\newcommand{\g}{|}
\newcommand{\EIG}{\mathrm{EIG}}
\newcommand{\sCEE}{\mathrm{sCEE}}

\newcommand{\entropy}{\mathrm{H}}

\DeclareRobustCommand{\KL}[2]{\ensuremath{\textsc{kl}\left[#1\;\|\;#2\right]}}

\DeclareMathOperator*{\argmax}{arg\,max}

\newcommand{\cA}{\mathcal{A}}
\newcommand{\cB}{\mathcal{B}}
\newcommand{\cD}{\mathcal{D}}

\newcommand{\cF}{\mathcal{F}}
\newcommand{\cG}{\mathcal{G}}

\newcommand{\cM}{\mathcal{M}}
\newcommand{\cN}{\mathcal{N}}

\newcommand{\cR}{\mathcal{R}}
\newcommand{\cS}{\mathcal{S}}
\newcommand{\cT}{\mathcal{T}}

\newcommand{\E}{\mathbb{E}}

\newacronym{DL}{dl}{deep learning}
\newacronym{VI}{VI}{variational inference}
\newacronym{BO}{BO}{Bayesian optimization}
\newacronym{DOE}{DoE}{design of experiments}
\newacronym{SGD}{SGD}{stochastic gradient descent}
\newacronym{EIG}{EIG}{expected information gain}
\newacronym{BOED}{BOED}{Bayesian optimal experimental design}
\newacronym[longplural=Markov decision processes]{MDP}{MDP}{Markov decision process}
\newacronym{RL}{RL}{reinforcement learning}
\newacronym{PCE}{PCE}{prior contrastive estimation}
\newacronym{SMC}{SMC}{sequential Monte Carlo}
\newacronym{SPCE}{sPCE}{sequential prior contrastive estimator}
\newacronym{SCEE}{sCEE}{sequential cross-entropy estimator}
\newacronym{DAD}{DAD}{deep adaptive design}
\newacronym{SED}{SED}{sequential experimental design}
\newacronym{REDQ}{REDQ}{Randomized ensembled double q-learning}
\newacronym{SNMC}{sNMC}{sequential nested monte carlo}
\newacronym{CES}{CES}{constant elasticity of substitution}
\newacronym{hipmdp}{HIP-MDP}{Hidden Parameter MDP}
\newacronym{SEDMDP}{SED-MDP}{sequential experiment design MDP}
\newacronym{SBI}{SBI}{simulation-based inference}
\newacronym{ABC}{ABC}{approximate Bayesian computation}

\newcommand{\BlackBox}{\rule{1.5ex}{1.5ex}}  % end of proof
\newenvironment{proof}{\par\noindent{\bf Proof\ }}{\hfill\BlackBox\\[2mm]}
 
\newtheorem{theorem}{Theorem}

\newtheorem{corollary}[theorem]{Corollary}

\icmltitlerunning{Statistically Efficient Bayesian Sequential Experiment Design via Reinforcement Learning with Cross-Entropy Estimators}

\begin{document}

\twocolumn[

%\icmltitle{Cross-Entropy Estimators for Sequential Experiment Design with Reinforcement Learning}
\icmltitle{Statistically Efficient Bayesian Sequential Experiment Design via Reinforcement Learning with Cross-Entropy Estimators}

\icmlsetsymbol{equal}{*}

\begin{icmlauthorlist}
\icmlauthor{Tom Blau}{nour,equal}
\icmlauthor{Iadine Chades}{env}
\icmlauthor{Amir Dezfouli}{bim,equal}
\icmlauthor{Daniel Steinberg}{d61}
\icmlauthor{Edwin V. Bonilla}{d61}
\end{icmlauthorlist}

\icmlaffiliation{nour}{Nourish Ingredients, Sydney, Australia}
\icmlaffiliation{env}{CSIRO's Environment, Brisbane, Australia}
\icmlaffiliation{bim}{BIMLOGIQ, Sydney, Australia}
\icmlaffiliation{d61}{CSIRO's Data61, Canberra \& Sydney, Australia}

\icmlcorrespondingauthor{Edwin V. Bonilla}{edwin.bonilla@csiro.au}

\icmlkeywords{Design of experiments, Reinforcement Learning}

\vskip 0.3in
]

\printAffiliationsAndNotice{\icmlEqualContribution} % otherwise use the standard text.

\begin{abstract}%
    Reinforcement learning can learn amortised design policies for designing sequences of experiments. However, current amortised methods rely on  estimators of expected information gain (EIG) that require an exponential number of  samples on the magnitude of the EIG to achieve an unbiased estimation. We propose the use of an alternative estimator based on the cross-entropy of the joint model distribution and a flexible proposal distribution. This proposal distribution approximates the true posterior of the model parameters given the experimental history and the design policy.  Our method overcomes the exponential-sample complexity of previous approaches 
    and provide more accurate estimates of high EIG values. More importantly, it allows learning of superior design policies, and is compatible with continuous and discrete design spaces, non-differentiable likelihoods and even implicit probabilistic models. 
\end{abstract}

\section{Introduction}

A key challenge in science is to develop predictive models based on experimental observations. As far back as~\citet{lindley1956measure} it has been recognised that experimental designs can be optimised to be maximally informative, under the assumptions of a Bayesian framework. Since then optimal experimental design has been applied to a wide variety of fields with different models and assumptions, including neuroscience~\citep{shababo2013bayesian}, biology~\citep{treloar2022deep}, ecology~\citep{drovandi2014sequential} and causal structure learning~\citep{agrawal2019abcd}.

Under the framework of \gls{BOED}, we have a probabilistic model $p(y \g \theta, d)$ where $d$ is the design (e.g.~where to measure), $y$ is the outcome (e.g.~the measurement value) and $\theta$ are parameters over which we have a prior belief $p(\theta)$. 
The objective is to find the optimal design $d^*$ that maximises the \gls{EIG}, defined as:
\begin{gather}
    \label{eq:first-eig}
    \EIG(d) \coloneqq \E_{p(y \g d)}\!\left[\entropy(p(\theta)) - \entropy(p(\theta \g y, d ))\right],\\
        \label{eq:first-optimal-d-eig}
    d^* = \argmax_{d \in \mathcal{D}} \EIG(d).
\end{gather}
Where $\entropy(\cdot)$ is the entropy of the given distribution. We see that na\"ive computation of the \gls{EIG} requires an expectation over the marginal likelihood $p(y \g d)$ and estimation of the posterior $p(\theta \g y, d)$. Since sampling from $p(y \g d)$ is typically intractable and there is usually no closed-form solution for the posterior, minimising this expression involves estimating a nested expectation numerically, which is challenging~\citep{rainforth2018nesting}. Furthermore, we are often interested in conducting more than one experiment, in which case optimal designs must incorporate the outcomes of previous experiments sequentially \citep{krause2007nonmyopic}.

In settings where computational or application-specific constraints demand fast deployment times, recent \textit{amortised} methods have proved successful, as they learn an optimal design policy as a function of the experimental history instead of optimising each design in turn~\citep{blau2022optimizing,foster2021deep,ivanova2021implicit}. Once trained, a policy can be reused to design experiments as many times as desired, thus amortising the cost of training. 
However, these methods have the drawback that they rely on maximising estimators of the \gls{EIG} that require an exponential number of samples on the magnitude of the \gls{EIG} to achieve an unbiased estimate \citep{mcallester2020formal, poole2019variational}.  Thus, their performance degrades in cases where the true \gls{EIG} is large. 
M
To address this limitation, we propose a new amortised method, using \gls{RL} and a bound based on the cross-entropy of the joint model distribution and a flexible proposal distribution. This bound can be seen as the sequential version of the bounds proposed by \citet{barber2004algorithm} and \citet{foster2019variational}. 
The proposal distribution approximates the true posterior of the model parameters given the experimental history and the design policy. Our method does not suffer from exponential sample complexity and is thus able to achieve higher \gls{EIG} than prior art, especially in settings where the information gain of the optimal policy is large. Furthermore, unlike previous amortised methods, our method is generally applicable to continuous and discrete design spaces, non-differentiable likelihoods, and even implicit likelihoods. 
Our experiments % provide supporting evidence of 
show the benefits of our approach when compared to competitive baselines.% We provide our code as supplementary material and will publish it upon acceptance.

% In the next section we cover the theoretical background necessary to understand our proposed method, which we explain in detail in~\cref{sec:theory,sec:method}. Experiments then follow in~\cref{sec:results} to test the claimed benefits of our approach. Related literature is discussed in~\cref{sec:related} and finally we present conclusions and potential future work in~\cref{sec:conclusion}.

\section{Amortised design of experiments}\label{sec:background}

In \acrfull{BOED} the goal is to identify the parameters of a probabilistic model by sending queries to that model. Let $p(y \g \theta, d)$ be the model of concern, with some prior belief $p(\theta)$ regarding the value of parameters $\theta$. 
As described above, an optimal design $d^*$ is one that maximises the \gls{EIG} as given by \cref{eq:first-eig,eq:first-optimal-d-eig}, where computational intractabilities readily appear in the estimation of the marginal likelihood and the posterior distribution. 

Furthermore, more challenging than optimising a single experiment is the problem of optimising an entire sequence of experiments $d_{1:T}$ where $T \in \mathbb{N}$ is some fixed budget. 
One promising approach for settings under strong computational constraints at deployment time is to optimise a design policy $\pi : \mathcal{H} \rightarrow \mathcal{D}$ that designs experiments conditioned on a history $h_t = (d_i, y_i)_{i=1}^t$. The computational cost of learning such a policy is high, but designing experiments with a trained policy is computationally efficient, requiring only a single forward pass of a neural network. Therefore the training cost is amortised over the lifetime of the policy, and this class of algorithms is known as \textit{amortised} sequential design of experiments.
% ~\citet{foster2021deep} have shown that the \gls{EIG} of a design policy can be maximised through a lower bound called \gls{SPCE}:
% \\
% \begin{align}
%     EIG(\pi, T) \ge sPCE(\pi, T, L) =  \E_{p(\theta_{0:L}) p(h_T \g \theta_0, \pi)} \left[ \log \frac{p(h_T \g \theta_0, \pi)}{\frac{1}{L+1} \sum^L_{l=0} p(h_T \g \theta_l, \pi)} \right], \label{eq:spce}
% \end{align}
% where $\theta_{1:L} \sim p(\theta)$ are \textit{contrastive samples}, giving the bound its name, and it is assumed that the probability of the history factorises into $p(h_T \g \theta, \pi) 
%  = \prod_{t=1}^T p(y_t \g \theta, d_t)$. A corresponding upper bound, sequential nested Monte Carlo (sNMC) estimator, can be achieved by excluding the $l=0$ term from the denominator.

A recent amortised method  proposed by \citet{foster2021deep} optimises the so-called \gls{SPCE}:
\begin{equation}
    \text{sPCE}(\pi, T) :=  \E \left[ \log \frac{p(h_T \g \theta_0, \pi)}{\frac{1}{L+1} \sum^L_{l=0} p(h_T \g \theta_l, \pi)} \right], \label{eq:spce}
 \end{equation}
where the expectation is taken over the joint ${p(\theta_{0:L}) p(h_T \g \theta_0, \pi)}$. It is easy to show that this estimator is a lower bound on the \gls{EIG}. Moreover, we see that it requires an original sample $\theta_0 \sim p(\theta)$ and additional samples $\theta_{1:L}$ in the denominator to compute the nested expectations. These additional samples are crucial to the construction of the bounded estimator \footnote{\citet{foster2021deep} refer to these $\theta_{1:L}$ samples as \emph{contrastive} as they can be seen as contrasts to the original sample  $\theta_0$. Generally, contrastive samples are used in various problems in machine  learning but notably in efficient learning of energy-based models (EBM). The basic idea is that one can learn EBMs by contrasting it with another distribution with known density \citep[see, e.g.,][]{pmlr-v9-gutmann10a}.}.

 Unfortunately, it has been shown that these types of bounds require a number of  samples $L$ that is exponential in the magnitude of the quantity being estimated~\citep{mcallester2020formal}. In other words, if the \gls{EIG} of a policy is large, then computing an accurate contrastive bound is intractable.
We emphasise here that the sample complexity of these estimators is exponential in the magnitude of the \gls{EIG}, rather than in the number of dimensions. 
This is usually referred to as the ``log-n curse'' or ``log-k curse'' rather than ``curse of dimensionality'' \citep[see, e.g,][]{chen-flatnce-2021,wang-smooth-infonce-2022}.

\section{The sequential cross-entropy estimator}\label{sec:theory}

% From Theorem 2 of~\citet{foster2021deep}, we have that the total EIG of a sequence of $T$ experiments, given a design policy, is lower bounded by the \gls{SPCE}. However, since the term $p(h_T \g \theta_0, \pi)$ appears in both the numerator and denominator of \gls{SPCE} (as per~\cref{eq:spce}), the lower bound is itself upper bounded by $\log(L+1)$, where $L$ is the number of contrastive samples. Therefore, to provide an unbiased estimate of $EIG(\pi, T)$, we require an exponential number of contrastive samples $L \ge \exp(EIG(\pi, T)) - 1$. 
As described above, the limitation of requiring an exponentially large number of samples inherent to contrastive EIG estimators is exacerbated significantly in the \textit{sequential} experimental design setting.  To address this, we propose the use of a cross-entropy estimator. To recap, our joint model distribution over parameters $\theta$ and observation history $h_T$ given a policy $\pi$ guiding the selection of designs is
\begin{align}
    \nonumber
    p(\theta, h_T \g \pi)&= p(\theta) \prod_{t=1}^{T} p(d_t \g \pi(y_{1:t-1}, d_{1:t-1})) p(y_t \g \theta, d_t) \\
    %p(\theta) p(h_T \g \theta, \pi),  \text{ where } \\
   % p(h_T \g \theta, \pi)  &= \prod_{t=1}^T p(y_t \g \theta, d_t),
   &= p(\theta) \prod_{t=1}^{T} p(h_t \g \theta, \pi(h_{t-1})), 
\end{align}
where $d_t = \pi(h_{t-1})$  and $h_0 \coloneqq \varnothing$.  For the above model we have that the true \gls{EIG} is given by (see \cref{app:bound_proof}):
    \begin{equation}
        % \nonumber
        % \EIG(\pi, T) = \E_{p(\theta, h_T \g \pi)} \left[ \log p(h_T \g \theta, \pi) - \log p(h_T \g \pi) \right],
       EIG(\pi, T) = \E_{p(\theta, h_T, \g \pi)} \left[ \log p(\theta \g h_T, \pi) \right] + \entropy\!\left[ p(\theta)
        \right],
    \end{equation}
which is, essentially, the sequential version of \cref{eq:first-eig} and, therefore, poses significant computational challenges.  
% where we note that this requires a marginalisation operation to obtain $p(h_T \g \pi)$ and an expectation over the full joint distribution $p(h_T, \theta \g \pi)$. 
% 
Our goal is then to obtain a tractable estimator of the \gls{EIG} that avoids the exponentially large number of samples required by contrastive approaches. For this purpose, we introduce a proposal distribution $q(\theta \g h_T, \pi)$ that approximates the true intractable posterior $p(\theta \g h_T, \pi)$. %In the sequential setting, this posterior is a function of a history of designs and observations, and a policy guiding the selection of designs. 
Furthermore, we propose using the following estimator that relies on the cross-entropy of these two distributions, and we refer to it as the \gls{SCEE}:
\begin{equation}
        \label{eq:sceet}
    \sCEE(\pi, T) \coloneqq \E_{p(\theta, h_T \g \pi)}\!\left[ \log q(\theta \g h_T, \pi) \right] + \entropy\!\left[ p(\theta) \right].
\end{equation}
From Jensen's inequality, it follows that the cross-entropy of two random variables is a lower bound for the self-entropy of either variable. By extending this to the sequential case, the following theorem shows that the sCEE is a lower bound of the true EIG:

\begin{theorem}\label{thm:bound}
    Let $p(y \g \theta, d)$ be a probabilistic model with prior $p(\theta)$. For an arbitrary fixed design policy $\pi$ and sequence length $T$, the \gls{EIG} of using $\pi$ to design $T$ experiments is denoted $\EIG(\pi, T)$.
    Let $q(\theta \g h_T, \pi)$ be a proposal distribution over parameters $\theta$ conditioned on experimental history $h_T$, and the sCEE bound as defined in~\cref{eq:sceet},
    % \begin{equation*}
    %     \sCEE(\pi, T) \coloneqq \E_{p(\theta, h_T \g \pi)} \left[ \log q(\theta \g h_T, \pi) \right] + \entropy\!\left[ p(\theta) \right],
    % \end{equation*}
    we have that
    \begin{equation*}
        \sCEE(\pi, T) \leq \EIG(\pi, T).
    \end{equation*}
\end{theorem}
\begin{proof}
    A sketch of the proof follows here, with the full proof in~\cref{app:bound_proof}. The main idea is to rewrite the \gls{EIG} as an expectation w.r.t.~distribution $p(h_T, \theta \g \pi)$, and then show that the difference between \gls{EIG} and \gls{SCEE} is an expectation over KL divergences.
\end{proof}
%
% ~\Cref{thm:bound} leads to $2$ corollaries that describe the nature of the lower bound (cf.~\cref{app:corollaries}):
% \begin{corollary}
%     The bound is tight if and only if $q(\theta \g h_T, \pi) = p(\theta \g h_T, \pi)$
% \end{corollary}
% \begin{corollary}
% The bias of the \gls{SCEE} estimator is $-\E_{h_T} \left[ \KL{p(\theta \g h_T, \pi)}{q(\theta \g h_T, \pi)} \right]$
% \end{corollary}
% In other words, the quality of the estimation rests on how well the proposal distribution can match the true posterior, in terms of KL divergence.
An important result follows from the above theorem describing the nature of the lower bound.
\begin{corollary}
The above bound is tight if and only if $q(\theta \g h_T, \pi) = p(\theta \g h_T, \pi)$.  The bias of the \gls{SCEE} estimator is $-\E_{h_T} \left[ \KL{p(\theta \g h_T, \pi)}{q(\theta \g h_T, \pi)} \right]$. 
\end{corollary}
This is straightforward to show (see \cref{app:corollaries}) and means that the quality of the estimation rests on how well the proposal distribution can match the true posterior, in terms of KL divergence. This compares remarkably favourably with  the bias of other estimators such as \gls{SPCE}, which has a bias of  $|\text{EIG}| - \log L$, where $L$ is the number of samples.

\textbf{Relation to previous bounds}: Our \gls{SCEE} bound described above is the sequential version of the bound proposed by \citet{barber2004algorithm}, who used it for estimating mutual information in the context of information transmission over noisy channels. This bound is also referred to in  
\citet{foster2019variational} as the variational posterior estimator, who used it for gradient-based experimental design in a non-sequential setting.
\subsection{Proposal parameterization}
To evaluate the \gls{SCEE}, we sample from the joint $p(\theta, h_T \g \pi)$ simply by rolling out the policy.  Under mild assumptions, it can be shown that this Monte Carlo estimation approaches the true value of the sCEE at a rate of $O(\frac{1}{\sqrt{n}})$, where $n$ is the number of samples (cf.~\cref{app:convergence_proof}). 
We will parameterise the proposal distribution by a conditional normalising flow network~\citep{winkler2019learning} with parameters $\kappa$ and, therefore, refer to it using $q_\kappa(\cdot)$. Thus, we can maximise the sCEE w.r.t.~$\kappa$ using stochastic gradient descent. Note that we only need to optimise the  
negative cross-entropy term $\E_{p(\theta, h_T \g \pi)}\!\left[ \log q_\kappa(\theta \g h_T, \pi) \right]$, since the prior entropy is constant. Details of the normalising flows used in our experiments are in~\cref{app:algos}.

\section{Experiment design with sCEE and reinforcement learning}\label{sec:method}
% To use the \gls{SCEE} bound in the context of an RL algorithm, we substitute sCEE for sPCE in the reward definition of the SED-MDP:
\citet{blau2022optimizing} have shown that the problem of learning an experiment design policy can be formulated as a special case of a \gls{MDP}, which we will refer to as \gls{SEDMDP}. Thus, we implement our proposed sequential design of experiments method by using the \gls{SCEE} bound in the formulation of the \textit{reward function} within the reinforcement learning (RL) framework defined by \citet{blau2022optimizing},%
\begin{align}
    \mathrm{R}(s_{t-1}, a_{t-1}, s_t, \theta) &= \log q_\kappa(\theta \g B_{\psi,t}, \pi_\phi) \nonumber \\
    &\quad- \log q_\kappa(\theta \g B_{\psi,t-1}, \pi_\phi), \label{eq:reward}
\end{align}%
where experimental designs correspond to policy actions $a_{t-1} = d_t$, and we have defined our proposal (approximate posterior) distribution as $q(\theta \g h_t, \pi) \coloneqq q_\kappa(\theta \g B_{\psi, t}, \pi_\phi)$. Here $B_{\psi,t} = \sum_{i=1}^t \mathrm{ENC}_\psi(y_i, d_i)$ maps history information to the system states, $s_t$, with a pooled summary from an encoder network  $\mathrm{ENC}_\psi (\cdot, \cdot)$\footnote{We note that this type of encoder was also used by \citet{foster2021deep} and \citet{blau2022optimizing}.}.

Furthermore, the parameterisation of history given by $B_{\psi,t}$ yields a permutation invariant pooled representation that is fed into the policy \emph{emitter} network\footnote{In practice we follow~\citet{blau2022optimizing} and learn a stochastic policy network that returns a distribution over designs, $\pi_\phi : \mathcal{H} \to \mathcal{P}_{\mathcal{D}}(\mathcal{D})$.}, $\pi_\phi(h_t) = \mathrm{EMM}_\phi(B_{\psi,t})$. This permutation invariance induces a Markovian structure, making such a parameterisation efficient for use in an RL framework. See~\cref{app:rl} for more details of the full formulation. Henceforth, we will refer to our method as RL-sCEE.
\subsection{The RL-sCEE Algorithm}
Policy and critic networks $\pi_\phi$ and $\mathcal{C}_\chi$ can be updated following the rules of the RL algorithm of our choice, using mini-batches of either off-policy or on-policy samples. In our experiments, we use the REDQ algorithm of \citet{chen2020randomized}. 
The posterior network $q_\kappa(\cdot)$ can be updated using the same mini-batches 
to maximise the log density of the observations under our posterior model. 
Note that this means rewards are now no longer fixed but depend on $q_\kappa(\cdot)$, and change with every update of the posterior parameters $\kappa$. The computational cost thus incurred can be minimised by lazy evaluation~\citep{bloss1988code}: we only update each reward when we are about to use it to update the policy and critic networks of the RL agent. The procedure is summarised in~\cref{alg:scee-rl}.

\begin{algorithm}[thb]
	\caption{RL-sCEE}
	\label{alg:scee-rl}
	\begin{algorithmic}
        \STATE {\bfseries Input:} $\mathcal{M}$: SED-MDP, $L_\pi$: policy loss function, $L_\mathcal{C}$: critic loss function. All as defined in \cref{app:rl}.
        \STATE Initialise replay buffer $\cB$
        \WHILE{convergence criterion not reached}
            \STATE Generate rollouts $(s_{0:T}, a_{0:T}, \theta)^{1:N}$ using $\cM$ and $\pi$ and push to $\cB$.
            \STATE Sample mini-batch of size $m$ from $\cB$
            \STATE Compute posterior loss\\ $L_q = -\frac{1}{m}\sum_{i=1}^{m}\log q_\kappa(\theta^i \g B_{\psi,t}^i)$
            \STATE Take gradient step to minimise $\nabla_\kappa L_q$
            \STATE Compute rewards for mini-batch using~\cref{eq:reward}
            \STATE Use mini-batch and rewards to compute $L_\pi$ \& $L_\mathcal{C}$
            \STATE Take gradient step to minimise $\nabla_\phi L_\pi$ and $\nabla_\chi L_\mathcal{C}$
        \ENDWHILE
    \end{algorithmic}
    %\vspace{-1em}
\end{algorithm}

We propose to learn the parameterised design policy network $\pi_\phi$ and the proposal distribution $q_\kappa(\cdot)$ from data simultaneously. Since the reward function depends on $q_\kappa(\cdot)$, and the objective function of $q_\kappa(\cdot)$ in turn depends on $\pi_\phi$, this leads to inherent instability, similar to the ``deadly triad'' that is often observed in value-based reinforcement learning~\citep{van2018deep}. We therefore apply several stabilisation mechanisms to prevent the neural network estimators from diverging. Details of the stabilisation mechanisms and RL formulation are given in~\cref{app:rl}.

\subsection{Advantages of RL-sCEE}

Our method based on the sCEE lower bound and RL delivers a number of advantages. \textbf{(i) Better sample complexity}: it does not require the use of contrastive samples, and hence does not suffer from the exponential sample complexity issue of the sPCE bound. 
Thus, sCEE can more closely estimate \gls{EIG} when the true quantity is large, although estimation accuracy depends on learning a good
 posterior network $q_\kappa(\cdot)$. \textbf{(ii) Applicable to implicit models}: 
Furthermore, we see that the sCEE estimator, as defined in \cref{eq:sceet}, only requires sampling of the model distribution and avoids explicit log-likelihood computations $\log p(h_T \g \theta, \pi)$. This means that our method is compatible with implicit likelihood models where the likelihood is a black-box or intractable and, therefore, can only be sampled but not evaluated explicitly. 
Interestingly, the sCEE bound is closely related to the sACE bound introduced in the appendices of~\citet{foster2021deep}. We discuss this relationship in~\cref{app:sace}. \textbf{(iii) Suitable for continuous and discrete design spaces}:  
Finally, similar to the method proposed in 
\citet{blau2022optimizing}, our approach  using the sCEE estimator along with reinforcement learning, as described in \cref{alg:scee-rl} (in~\cref{app:rl}), can handle both continuous and discrete design spaces.

\section{Experimental results}\label{sec:results}

We evaluate our proposed method on (i) synthetic data; (ii) continuous designs and implicit likelihoods\footnote{We simulate an implicit likelihood by withholding the explicit likelihood values $p(y \g \theta, d)$ from the RL-sCEE and iDAD agents.} in behavioural economics under a constant elasticity of substitution (CES) problem and a (iii) source location problem; and (iv) discrete designs in a prey population problem. Description and details of these problems and their mathematical models are given in the subsequent subsections.   
We compare our method (RL-sCEE) to a number of baselines; RL with the sPCE bound \citep[RL-sPCE;][]{blau2022optimizing}, Deep Adaptive Design \citep[DAD;][]{foster2021deep}, implicit Deep Adaptive Design \citep[iDAD;][]{ivanova2021implicit}, and a non-amortised sequential Monte Carlo experiment design approach \citep[SMC-ED;][]{moffat2020sequential}. We also compare to two greedy baselines; one that maximises the variational PCE bound~\citep[VPCE;][]{foster2020unified}, and one that samples designs uniformly at random (Random).

\subsection{Accuracy of estimators on synthetic data}

\begin{table*}[tbh]
\caption{Different estimators for EIG of increasing magnitudes in synthetic data problems with conjugate priors. Averages computed over $1000$ samples. $k$ is the number of random variable dimensions, $\sigma_0$ is prior variance, and $\sigma$ is likelihood variance.}
\label{table:estimators}
%\vspace{-.8em}
\begin{center}
\begin{small}
\begin{tabular}{lccccccc}
\toprule
Method & $k=10$ & $k=10$ & $k=10$ & $k=10$ & $k=10$ & $k=10$ & $k=20$ \\
    & $\sigma_0 = 0.5$ & $\sigma_0 = 0.5$ & $\sigma_0 = 1$ & $\sigma_0 = 2$ & $\sigma_0 = 2$ & $\sigma_0 = 4$ & $\sigma_0 = 4$ \\
    & $\sigma = 5$ & $\sigma = 1$ & $\sigma = 1$ & $\sigma = 1$ & $\sigma = 0.5$ & $\sigma = 0.5$ & $\sigma = 0.5$ \\
\midrule
True EIG & $3.47$ & $8.96$ & $11.99$ & $15.22$ & $18.57$ & $21.97$ & $43.94$ \\
sCEE & $3.40$ & $\mathbf{8.90}$ & $\mathbf{11.92}$ & $\mathbf{15.07}$ & $\mathbf{18.41}$ & $\mathbf{20.47}$ & $\mathbf{43.89}$ \\
sPCE($L=1\textrm{E}4)$ & $\mathbf{3.45}$ & $7.92$ & $8.95$ & $9.18$ & $9.21$ & $9.21$ & $9.21$  \\
sPCE($L=1\textrm{E}6)$ & $3.48$ & $8.89$ & $11.45$ & $13.18$ & $13.75$ & $13.81$ & $13.81$ \\
sPCE($L=1\textrm{E}8)$ & $3.48$ & $8.97$ & $11.85$ & $14.35$ & $16.71$ & $18.08$ & $18.42$  \\
\bottomrule
\end{tabular}
\end{small}
\end{center}
\vspace{-.5em}
\end{table*}

Given the theoretical results about sCEE and contrastive bounds, we expect that sCEE should perform well in situations where the \gls{EIG} is large and $q_\kappa(\cdot)$ is easy to learn. To assess this, we evaluate the estimator on $7$ experimental tasks which allow us to know the true \gls{EIG} in closed form. The priors are isotropic Gaussians of the form $\cN(\mu_0, \sigma_0\mathbf{I}_k)$ and the likelihoods are similarly Gaussian with known isotropic covariance $\sigma\mathbf{I}_k$, where $k$ is the number of dimensions. Each task has an experimental budget of $T=10$ experiments. Thus we can manipulate $k, \sigma_0$ and $\sigma$ to create tasks where the \gls{EIG} of the optimal design is known exactly. See~\cref{app:conjugate} for details.

\Cref{table:estimators} enumerates these tasks, alongside the optimal \gls{EIG} and the estimates of sCEE and sPCE with different numbers of contrastive samples. As can be seen from the left-most columns of the table, when the \gls{EIG} is small enough, sPCE can provide a better estimate than sCEE (note that the sPCE at times slightly overestimates the \gls{EIG} due to variance in estimating the expectation with Monte Carlo samples). However, as the \gls{EIG} becomes large relative to $\log(L)$, the underestimation of sPCE becomes more severe, and for the right-most columns all sPCE variants have reached their upper limit. Meanwhile, sCEE consistently provides good estimates regardless of the magnitude of the \gls{EIG}. It should be noted, however, that this is in part because in this experiment the posterior is easy to learn from data. We will evaluate more complex posteriors in the following sections.
%A more complex posterior, or less training, would worsen the underestimation.
%
%
\subsection{Continuous designs}
\begin{figure}[htb]
\centering
\subfloat[CES experiment\label{fig:ces_exp}]{\includegraphics[width=0.7\columnwidth]{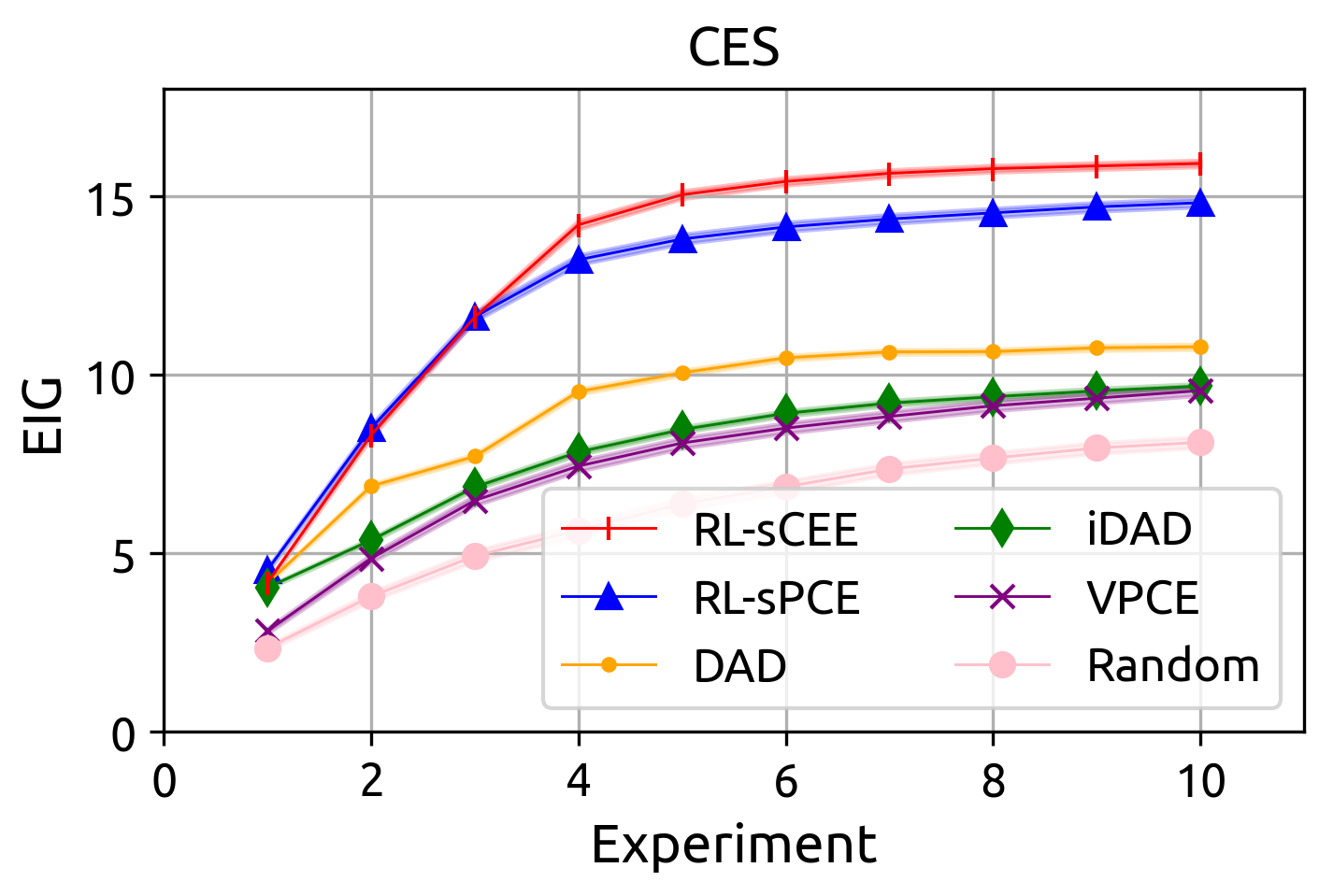}}\\
%\vspace{-.7em} \\
\subfloat[Source experiment\label{fig:source_exp}]{\includegraphics[width=0.7\columnwidth]{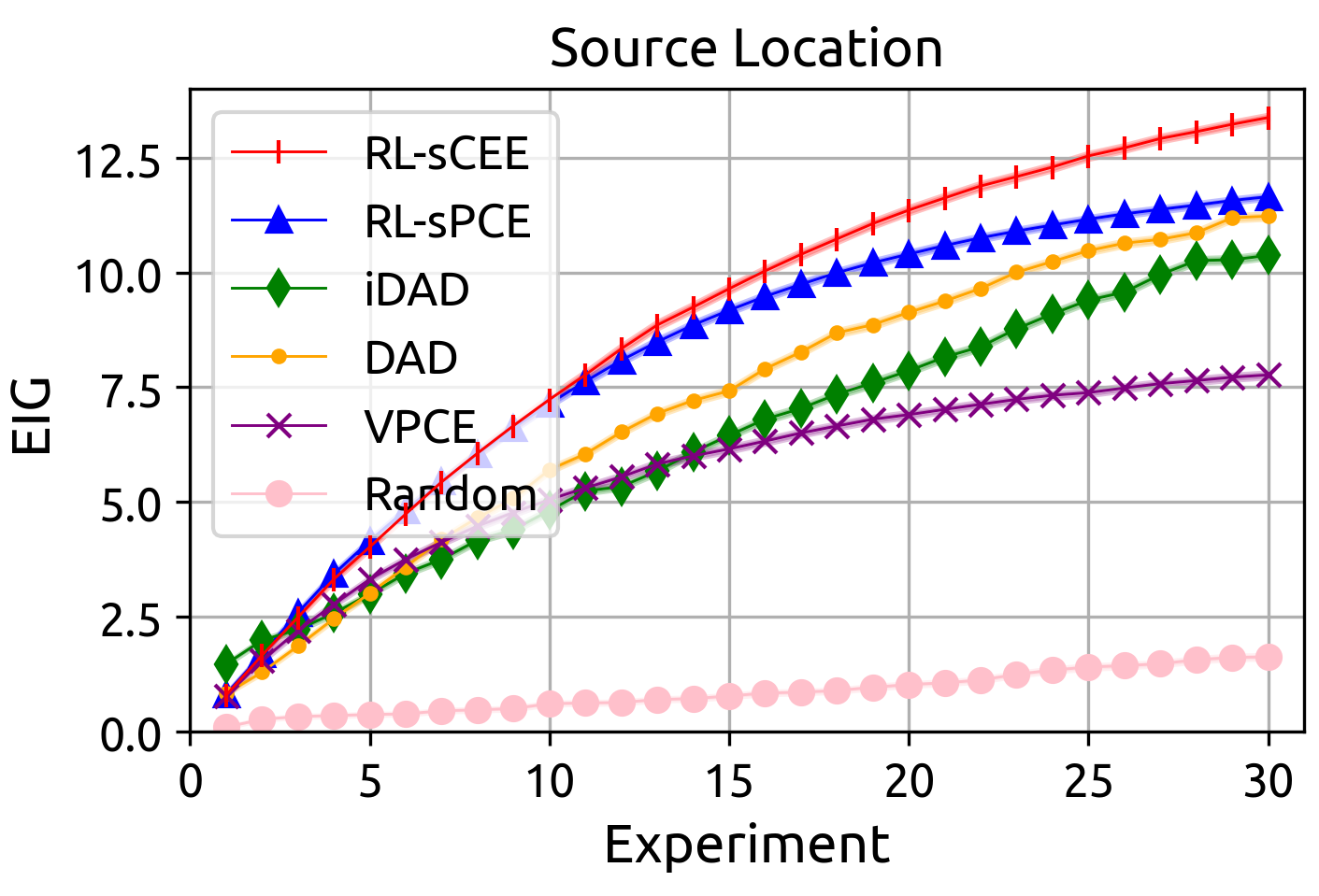}}
%\vspace{-.3em}
\caption{\gls{EIG} for the CES  % ~\cref{fig:ces_exp} 
and the source location problems, %~\cref{fig:source_exp}, 
estimated using \gls{SPCE} with $L=1\textrm{E}8$. Trendlines are means and shaded regions are standard errors aggregated from $1000$ rollouts. Our method is referred to as RL-sCEE.}
\label{fig:ces_source}
\vspace{-1.5em}
\end{figure}
\begin{figure}[htb]
\begin{center}
\centerline{
    \includegraphics[width=1.2\columnwidth]{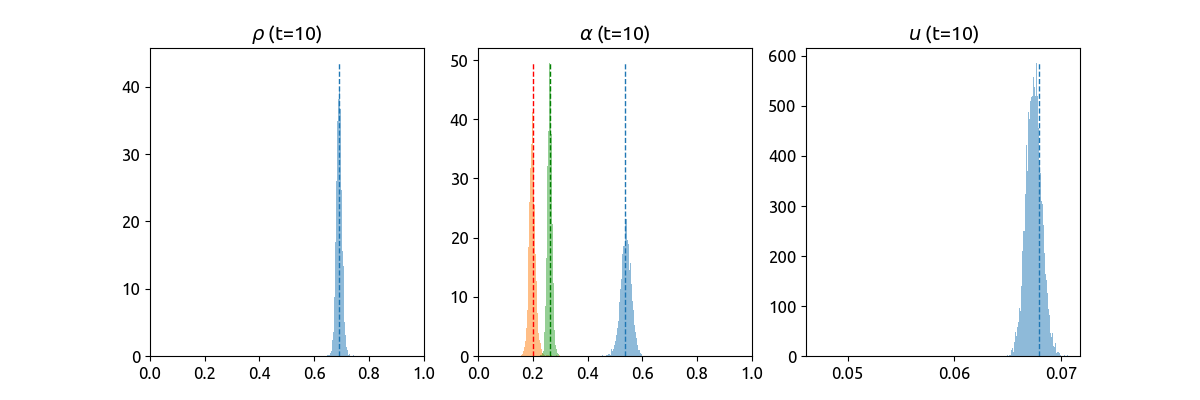}
}
\vspace{-.5em}
\caption{Example posterior distributions for the CES problem after $10$ experiments. Histograms are based on $1\textrm{E}5$ samples. Dashed vertical lines indicate the ground truth value of each variable. The middle plot shows the marginals of the $3$ different elements of $\alpha$.}
\label{fig:post_ces}
\end{center}
\vspace{-2.5em}
\end{figure}
\begin{figure}[tbh]
    \centering
    \subfloat[CES experiment \label{subfig:ces_prop}]
        {\includegraphics[width=0.75\columnwidth]{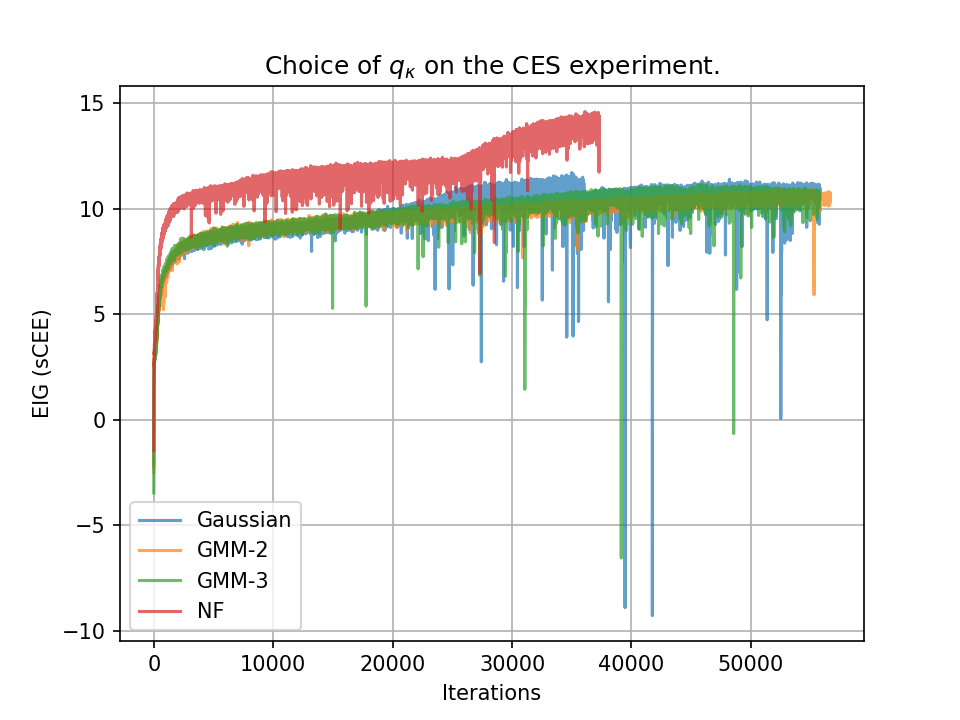}} \\
    \vspace{-1em}
    \subfloat[Source experiment \label{subfig:source_prop}]
        {\includegraphics[width=0.75\columnwidth]{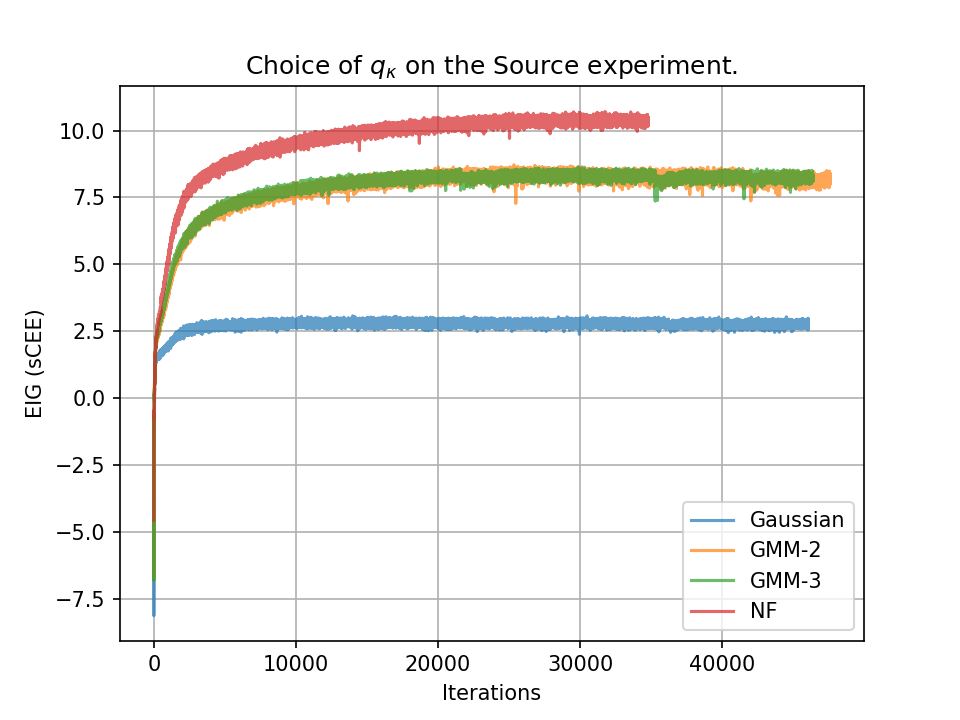}}
    \vspace{-.3em}
    \caption{Influence of the choice of proposal distribution, $q_\kappa(\cdot)$, on the sCEE reward for a set wall-time limit. \cref{subfig:ces_prop} shows the CES experiment with $T=10$ with a 72 hour wall-time limit. \cref{subfig:source_prop} shows the Source experiment with $d=2$ and $T=30$ with a 60 hour wall-time limit. The proposal distributions are normalising flows (NF), Gaussian, and Gaussian mixtures with two and three components (GMM-2, GMM-3).}
    \label{fig:prop_ablation}
    \vspace{-1.5em}
\end{figure}
\begin{figure*}[htb]
\begin{center}
\centerline{
    \includegraphics[width=0.7\linewidth]{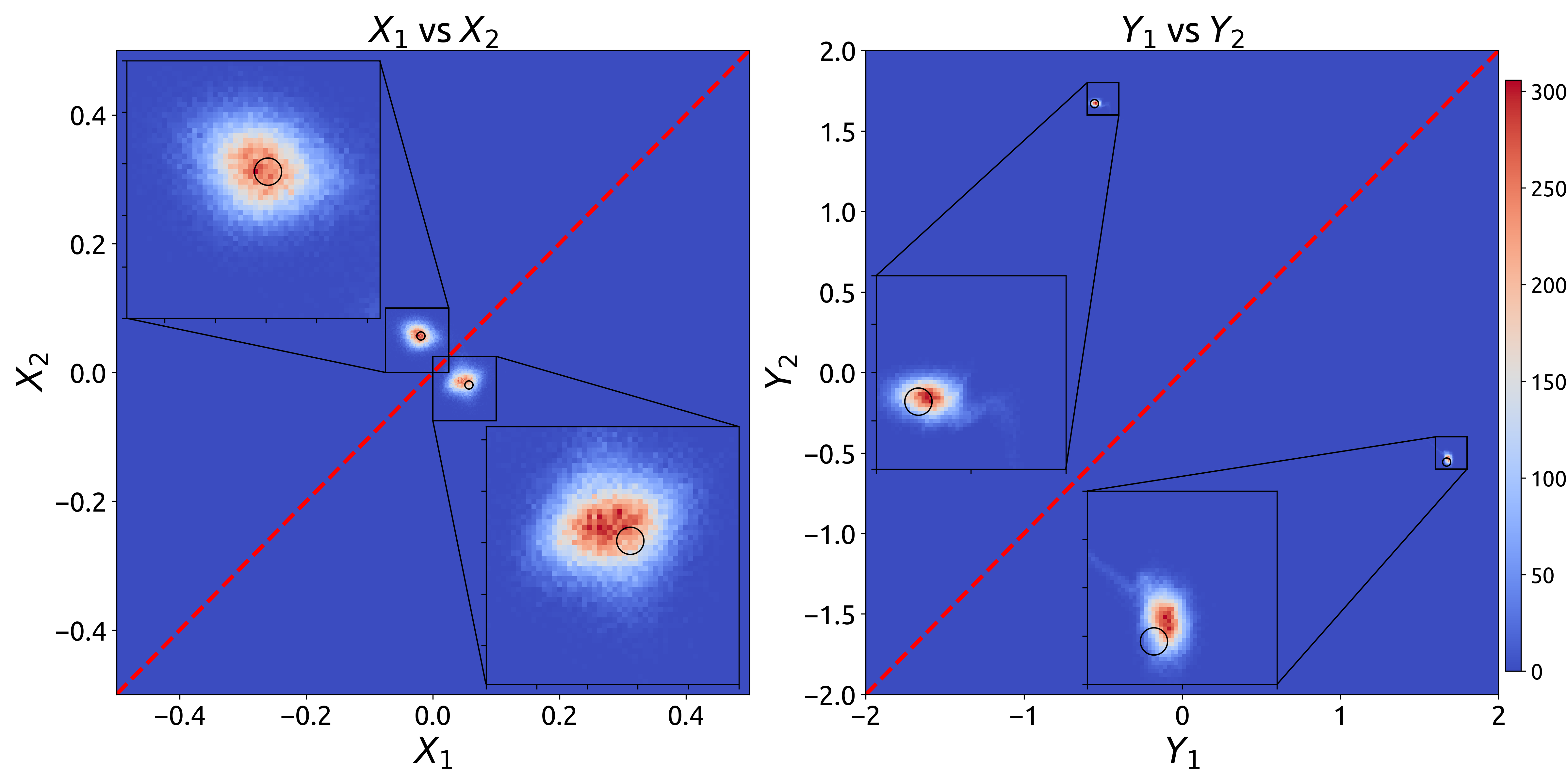}
}
\vspace{-1.5em}
\caption{Histograms of example posteriors for the source location problem after $30$ experiments, showing the joint distributions of the $x$ co-ordinates (left) and $y$ co-ordinates (right) of the $2$ sources. The plots show symmetry with respect to the dashed red line, which is predicted by Bayes' theorem. Insets zoom in on the modes of each posterior. Histograms are based on $1\textrm{E}5$ samples. Black rings denote the ground truth value of each variable.}
\label{fig:post_source}
\vspace{-2em}
\end{center}
\end{figure*}
\textbf{Constant elasticity of substitution (CES)}: Moving on to realistic tasks, we evaluate a design problem in behavioural economics where we must estimate the parameters of a Constant Elasticity of Substitution (CES) utility function~\cite{baltas2001utility}. Experimental designs consist of $2$ bags of goods with $3$ goods in each, so that the design space is $\cD = [0,100]^6$. The outcome is the relative preference of a test subject in the range $[0,1]$, as determined by the agent's CES utility function, and the specific values of its parameters $\theta = \{\rho, \alpha, u\}$, with $\rho \in [0,1]$, $ \alpha \in \Delta_3$ and  $u > 0$. See~\cref{app:ces} for full details.

\Cref{fig:ces_exp} shows the \gls{EIG} achieved at each point in a sequence of $T=10$ experiments. \gls{EIG} was estimated by the sPCE estimator with $L=1\textrm{E}8$ contrastive samples. Our proposed method, RL-sCEE, performs significantly better than all baselines. Indeed, by the $5^{th}$ experiment, our method already exceeds the performance that the state-of-the-art RL-sPCE baseline attains after $10$ experiments. Considering the sNMC upper bound shown in~\cref{table:eig}, it is clear that the \textit{lower} bound for RL-sCEE is higher than the \textit{upper} bound for RL-sPCE. This is particularly important since the \gls{EIG} for the RL-sCEE policy is approaching the limits of what can be estimated by the lower bound, so that the gap between the true \gls{EIG} and the estimate is potentially large.

Since the conditional normalising flow $q_\kappa$ is a posterior distribution, we can efficiently sample from it for any given history. It is important to note, however, that such posteriors can be expected to be accurate only for histories sampled from the joint $p(\theta, \pi_\phi)$ where $\pi_\phi$ is the stochastic policy network that was trained alongside $q_\kappa(\cdot)$. A representative posterior is shown in~\cref{fig:post_ces}. The histograms were computed using $1\textrm{E}5$ samples and $500$ bins. The marginal distribution for each random variable is unimodal and highly concentrated. The mode of each marginal is very close to the true value of the corresponding random variable, denoted by a dashed vertical line of the same colour.

\emph{Posterior ablation}: In order to test the efficacy of using the conditional normalising flow for the approximating the posteriors, $q_\kappa(\cdot)$, we compare it to using alternative parametrizations in~\cref{subfig:ces_prop}. We use a Gaussian, and Gaussian mixture distributions with 2 and 3 components as alternatives. For the same wall-time (72 hours), the conditional normalising flow obtains a higher \gls{SCEE} estimate of \gls{EIG} even though fewer learning iterations are performed.

\textbf{Source location}: In the next experiment we consider the problem of locating signal sources in space. Here we have $2$ sources with co-ordinates $s_1 = (x_1,y_1); s_2=(x_2,y_2)$, and designs consist of choosing the co-ordinates at which to take a noisy sample of signal magnitude. Signals decay with distance from the source and individual signals always superpose constructively. The full probabilistic model is described in~\cref{app:source}.

We trained policies to design sequences of $T=30$ experiments. The mean and standard error of \gls{EIG} achieved by our proposed method, as well as several baselines, are shown in~\cref{fig:source_exp}. RL-sCEE outperforms all baselines by a statistically significant margin. As before, ~\cref{table:eig} shows that the lower bound estimate for our proposed method exceeds the upper bound estimate for all other methods.

As with the CES problem, we can examine the posterior obtained from $q_\kappa(\theta \g h_T, \pi)$. This time, however, we plotted the marginals $q(x_1,x_2)$ and $q(y_1, y_2)$, i.e. the $x$ and $y$ co-ordinates of the $2$ signal sources, using $x_1$ (respectively $y_1$) as the X-axis and $x_2$ (respectively $y_2$) as the Y-axis. The sources are exchangeable, i.e. $p(y|d, s_1=(x_1,y_1), s_2=(x_2,y_2)) = p(y|d, s_1=(x_2,y_2), s_2=(x_1,y_1))$. Therefore the marginals of the true Bayesian posterior, $p(x_1,x_2)$, should be symmetric w.r.t. the line $x_2 = x_1$ in the $x_1x_2$-plane. Indeed, the plots in~\cref{fig:post_source} exhibit this symmetry, which has been learned entirely from data, without providing any inductive bias. We include a histogram of the full posterior in~\cref{app:results}.

\emph{Posterior ablation}: Again, we test different characterisations of the posterior, $q_\kappa(\cdot)$, to justify our choice of conditional normalising flows in~\cref{subfig:source_prop}. Our choice of conditional normalising flows performs better than all alternatives tested for estimating \gls{EIG} with \gls{SCEE} given a wall-time budget of 60 hours.
% EVB is here so far 

\emph{Dimensionality ablation}: Here we compare how the \gls{SCEE} and \gls{SPCE} reward functions operate in higher dimensions in~\cref{tab:dimensiont30}. This is a similar experiment as that conducted by~\citet{ivanova2021implicit} (their Table 2), but we use an experimental sequence of $T=30$. All estimators were run for a 60-hour wall time. % We found that for $d=\{6, 10\}$ \gls{EIG} was not always improved during training for both RL methods, and so we report on the models that achieved the highest estimated training \gls{EIG}.
We used the estimated training \gls{EIG} to track the best model, which for $d=\{6, 10\}$ often occurred before the maximum training budget for both RL methods.  

%We present extended results in~\cref{app:dimension} for other experimental horizons, $T$. 
We found that RL-sCEE required more iterations before it converged compared to RL-sPCE, however RL-sCEE has a smaller memory footprint and so can be parallelized more readily. In order to fit some of the experiments into GPU memory (16GB), we had to reduce several aspects of our experimental settings (detailed in the caption of~\cref{tab:dimensiont30}), therefore the numbers here are not comparable with those in~\cref{table:eig}.
%we had to lower the number of parallel tasks (n\_parallel) for \gls{SPCE}, which lowers the number of iterations for a given wall-time, and for both methods we had to reduce the buffer capacity from 1E7 to 1E6 for the $d=6$ and 10 runs. We have only run $L$ = 1E6 instead of 1E8 to evaluate the lower (\gls{SPCE}) and upper (\gls{SNMC}) bounds since the memory requirements become too large for more contrastive samples when $d > 2$. Because of this and different run-times, the upper and lower bounds are different from those reported in~\cref{table:eig}.

In lower dimensional settings, where \gls{EIG} is higher, the \gls{SCEE} method outperforms \gls{SPCE}. As we increase dimensionality \gls{SPCE} outperforms \gls{SCEE} perform similarly. This may be because \gls{SPCE} is better at estimating lower \gls{EIG}s as we saw in the synthetic data experiment.
%We also hypothesised that the conditional normalising flows proposal distribution, $q_\kappa(\cdot)$, used by the \gls{SCEE} method may need increasingly more computation or is more sensitive to initialisation as the dimensionality of $\theta$ increases. This would lead to worse performance of the \gls{SCEE} in higher dimensions. To test this, we ran the \gls{SCEE} method with a two-component Gaussian mixture (one for each source) proposal in the $d=10$ experiment. It clearly outperformed \gls{SCEE}-NF and \gls{SPCE}, suggesting choice of proposal distribution, $q_\kappa(\cdot)$, can have a strong influence on the performance of the method.

\textbf{Implicit likelihood}: in both the CES and source location problem, we simulate an implicit likelihood by withholding the explicit likelihood values $p(y \g \theta, d)$ from the RL-sCEE and iDAD agents. One would expect that both agents would perform worse than the RL-sPCE and DAD agents, which do exploit explicit likelihood information. Indeed, the iDAD agent is the least performant of any method considered. However, our proposed RL-sCEE method outperforms all baselines, both in the source location and CES problems, in spite of not having access to explicit likelihoods.

\begin{table*}[bth]
\caption{Lower and upper bounds for the EIG computed using the sPCE and sNMC estimators, respectively. $L=1\textrm{E}8$ contrastive samples were used for the CES and Source Location problems, and $L=1\textrm{E}6$ for the Prey Population problem. Means and standard errors aggregated from $1000$ rollouts.}
%\vspace{-.8em}
\label{table:eig}
\setlength\tabcolsep{4.75 pt}
\begin{center}
\begin{small}
\begin{tabular}{lcccccc}
\toprule
Method & \multicolumn{2}{c}{CES}  & \multicolumn{2}{c}{Source Location} & \multicolumn{2}{c}{Prey Population}  \\
    & Lower bound & Upper bound & Lower bound & Upper bound & Lower bound & Upper bound \\
\midrule
RL-sCEE & $\mathbf{15.91} \pm 0.10$ & $20.78 \pm 0.43$ & $\mathbf{13.37} \pm 0.07$ & $13.42 \pm 0.08$ & $4.41 \pm 0.05$ & $4.41 \pm 0.05$ \\
RL-sPCE & $14.81 \pm 0.12$ & $15.56 \pm 0.17$ & $11.65 \pm 0.06$ & $12.01 \pm 0.07$ & $4.38 \pm 0.05$ & $4.41 \pm 0.04$ \\
DAD & $10.77 \pm 0.08$ & $13.20 \pm 0.68$ & $11.22 \pm 0.07$ & $11.29 \pm 0.07$ & N/A & N/A \\
iDAD & $9.67 \pm 0.08$ & $10.63 \pm 0.52$ & $10.37 \pm 0.07$ & $10.41 \pm 0.08$ & N/A & N/A \\
SMC-ED & N/A & N/A & N/A & N/A & $\mathbf{4.52} \pm 0.07$ & $4.52 \pm 0.06$ \\
\bottomrule
\end{tabular}
\end{small}
\end{center}
\vspace{-1em}
\end{table*}

\begin{table}[tbh]
\vspace{-.5em}
\centering
\caption{Behaviour of the % RL-sCEE and RL-sPCE 
estimators on the Source experiment with $T=30$ in increasingly high dimensions. Lower and upper bounds for the EIG computed using the sPCE and sNMC estimators, respectively, with $L$ = 1E6 contrastive samples and 1000 rollouts. 
%All methods were run for a 60 hour wall-time.
% A buffer capacity of 1E7 was used for $d=\{2, 4\}$ but had to be reduced to 1E6 for $d=\{6, 10\}$ because of GPU memory limitations.
}
\vspace{.5em}
\begin{small}
\begin{tabular}{rlccccc}
\toprule
$d$ & Method & EIG lower & EIG upper \\
\midrule
2 & RL-sCEE & 12.31 ± 0.06 & 14.29 ± 0.16 \\
 & RL-sPCE & 12.13 ± 0.05 & 12.97 ± 0.08 \\
4 & RL-sCEE & 8.99 ± 0.08 & 9.09 ± 0.09 \\
 & RL-sPCE & 8.87 ± 0.08 & 8.98 ± 0.09 \\
6 & RL-sCEE & 4.37 ± 0.08 & 4.37 ± 0.08 \\
 & RL-sPCE & 4.69 ± 0.08 & 4.70 ± 0.08 \\
10 & RL-sCEE & 1.71 ± 0.05 & 1.71 ± 0.05 \\
% & RL-sCEE GMM-2 & 1.65 ± 0.05 & 1.65 ± 0.05 \\
 & RL-sPCE & 1.85 ± 0.05 & 1.85 ± 0.05 \\
\bottomrule
\end{tabular}
\end{small}
\vspace{-1.em}
\label{tab:dimensiont30}
\end{table}

\subsection{Discrete designs}

\begin{figure}[htb]
\centering
\subfloat[\label{fig:eig_prey}]{\includegraphics[width=0.75\columnwidth]{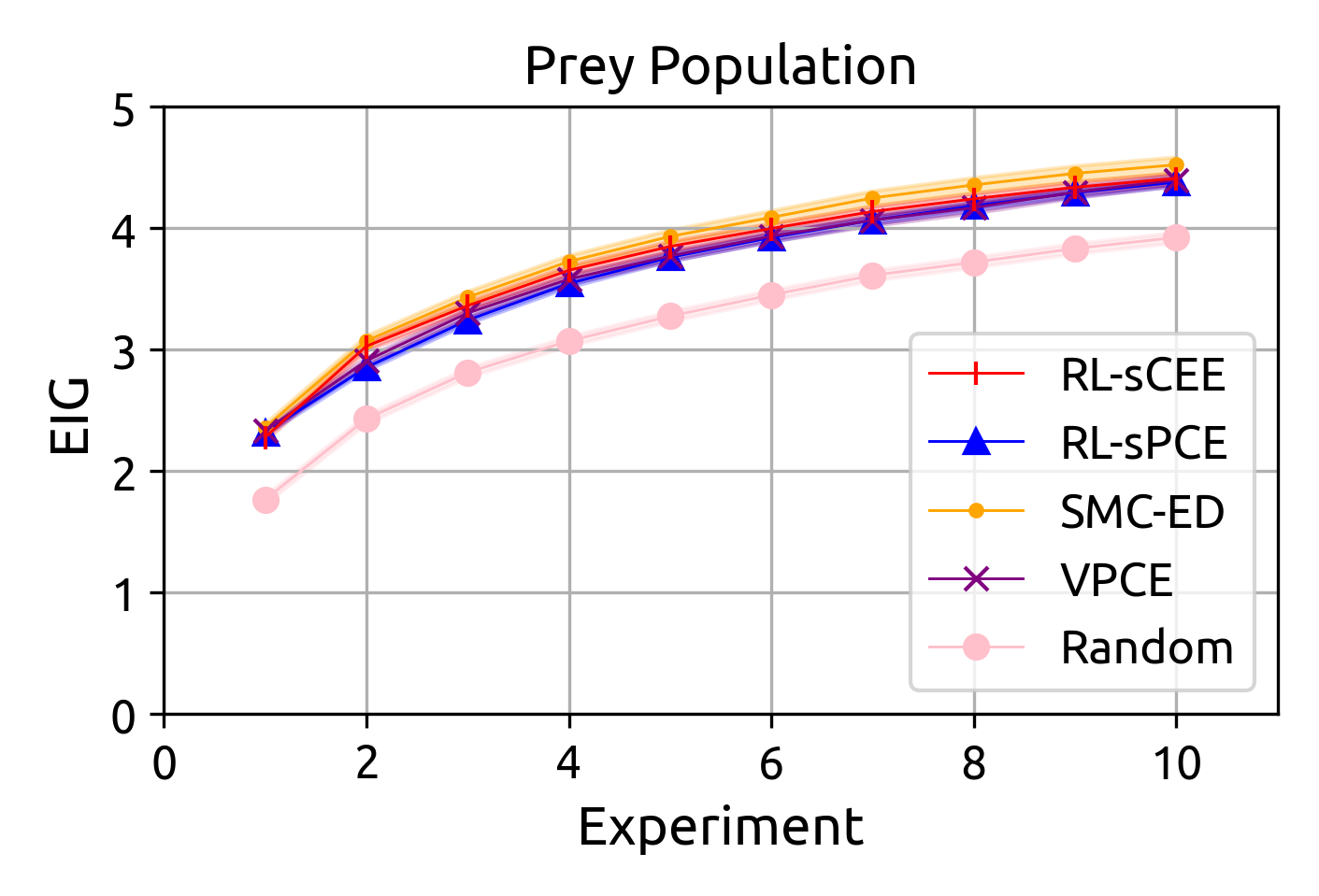}} \\
\vspace{-.7em}
\subfloat[\label{fig:dist_prey}]{\includegraphics[width=0.75\columnwidth]{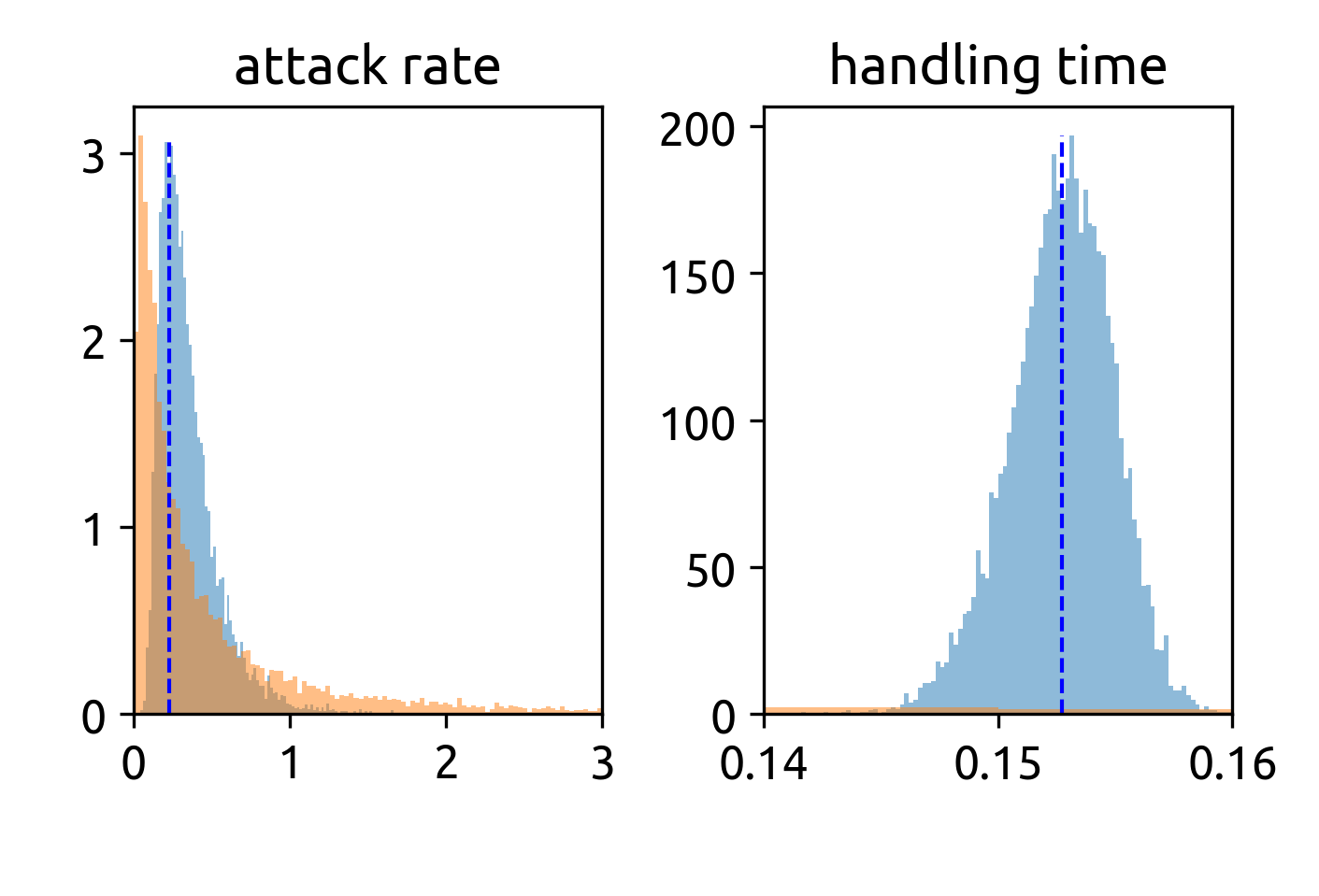}}
\vspace{-.3em}
\caption{\cref{fig:eig_prey} \gls{EIG} for the prey population problem, estimated using \gls{SPCE} with $L=1\textrm{E}6$. Trendlines are means and shaded regions are standard errors aggregated from $1000$ rollouts (\gls{RL}) or $500$ rollouts (SMC-ED). \cref{fig:dist_prey}  priors (orange) and posteriors (blue) after $10$ experiments. Histograms used $1\textrm{E}5$ samples.}
\label{fig:prey}
\vspace{-1em}
\end{figure}

To evaluate our method in tasks with discrete design spaces, we consider the prey population problem from~\citet{moffat2020sequential}. Designs are the initial population of a prey species, limited to the discrete interval $\cD = {1,2, \dots, 300}$. The outcome is the number of individual who were consumed by predators at the end of a $24$ hour period, based on the attack rate and handling time of the predators. Full details are available in~\cref{app:prey}.

Since DAD and iDAD cannot optimise over discrete design spaces, we added the sequential Monte Carlo design algorithm proposed by~\citet{moffat2020sequential} as a baseline. Note that this method is not amortised, and requires considerable computation time to design each experiment ($>1$ minute per design, whereas amortised policies take milliseconds). As can be seen from~\cref{fig:eig_prey}, RL-sCEE performs similarly to the baselines, in spite of having no access to explicit likelihood information (a circumstance which would cause both baselines to fail) and using orders of magnitude less time to compute designs than the SMC-ED baseline. The numerical estimates in~\cref{table:eig} show a relative difference of $\sim1\%$ in the mean \gls{EIG} estimates, and the standard errors overlap.

Finally, \cref{fig:dist_prey} shows the posterior distributions after $10$ experiments, with their corresponding priors. Both marginals place the mode near the true value of the random variable. The handling time (right) is fitted very tightly, whereas the attack rate (left) has a poorer fit. Both variables have the same prior, but the posterior for handling time is so concentrated that the prior is barely visible at the bottom of the plot.

\section{Related work}\label{sec:related}

Considerable work has been done on \gls{BOED}~\citep{chaloner1995bayesian, ryan-doe-review-2016}, and particularly on using machine learning to optimise experimental designs~\citep{rainforth2023modern}. Greedy algorithms have been developed based on variational bounds~\citep{foster2019variational,foster2020unified} or neural network estimates~\citep{kleinegesse2020bayesian} of the \gls{EIG}. In the active learning literature, the BALD~\citep{houlsby2011bayesian} score is equivalent to \gls{EIG}, and can be estimated using Monte Carlo dropout neural networks~\citep{gal2017deep}. Other works attempt a non-greedy approach, i.e. they can sacrifice information gain in the current experiment in exchange for higher information gain in future experiments. Such approaches include n-step look-ahead~\citep{zhao2021uncertainty,yue2020non} or using batch designs as a lower bound for the utility of sequential designs~\citep{jiang2020binoculars}.
~\citet{foster2021deep} were the first to propose an amortised method for sequential experiment design, and showed empirically that the learned policies can exhibit non-myopic behaviour. This was extended to the case of implicit likelihood models by~\citet{ivanova2021implicit}. ~\citet{blau2022optimizing} formulated the \gls{SED} problem as a special \gls{MDP}, and showed that design policies can be learned with \gls{RL} algorithms.

There is also a similarity (up to the distribution of the expectation) between \gls{SCEE} bound and the cross entropy objectives used in \gls{SBI} -- in particular the amortized posterior variants~\cite{cranmer2020frontier} that admit implicit treatments of the likelihood function in the style of \gls{ABC}. Furthermore, cross entropy methods have a rich history in optimisation and importance sampling for rare event estimation and conditioning~\cite{botev2013cross, rubinstein1999cross, miller2021rare, brookes2019conditioning}.

The field of Reinforcement Learning has two specialised frameworks for dealing with unknown parameters that govern the system dynamics: Bayes Adaptive MDPs~\citep{duff2002optimal} and Partially Observable MDPs~\citep{kaelbling1998planning}. In both cases, agents must balance between improving their posterior belief about the MDP and achieving high rewards. The variBAD~\citep{zintgraf2021varibad} algorithm fits a variational autoencoder to observed transitions and rewards, and its latent variables are used as input for the policy. The FORBES~\citep{chen2022flow} algorithm, on the other hand, uses normalising flows to model beliefs over latent variables, transitions and rewards. MAX~\citep{shyam2019model} uses an ensemble model to generate synthetic data and compute information gain, which guides each policy action.

Finally, as mentioned in \cref{sec:theory}, our \gls{SCEE} estimator is the sequential version of the bound proposed in \cite{barber2004algorithm} and the variational posterior estimator of \cite{foster2019variational}. Our work then focuses on its extension and usage in the sequential setting with RL and demonstrates its practical advantages with respect to previous amortised DOE methods, including that of~\cite{blau2022optimizing}.

\section{Conclusions \& limitations}\label{sec:conclusion}

We have introduced the sequential Cross-Entropy Estimator (\gls{SCEE}), a lower bound estimate for the \gls{EIG} of an experiment design policy, as well as a reinforcement learning algorithm (RL-sCEE) that uses it to optimise policies. %Experiments show 
%the sCEE is capable of estimating large \glspl{EIG} that are intractable to estimate with contrastive estimators, which are the state of the art. 
% In tasks where \gls{EIG} is large, RL-sCEE significantly outperforms all baselines and learns policies whose lower bound \gls{EIG} estimates exceed the upper bound estimate of the strongest alternative. In tasks where \gls{EIG} is small, RL-sCEE matches the performance of state-of-the-art baselines.
We have shown that RL-sCEE can outperform or be comparable to state-of-the-art baselines.

RL-sCEE relies on learning a parameterized proposal network, $q_\kappa(\cdot)$, that closely matches the true posterior. In problems where this is challenging, the estimator bias could be large, and the design policies will be degraded.
% Our experiments also show that with 
Furthermore, it is possible for a contrastive estimator to outdo \gls{SCEE} with an extremely large number of samples.  However, the cost of doing so during training, where millions of estimations are needed, is prohibitive for current hardware. Similarly, it is difficult to use the  \gls{SCEE} post-hoc for a policy that was not co-trained with the estimator, as this requires learning the policy network from scratch.
Future work can focus on improving the learning of this proposal network.
% , which will accelerate the training of design policies as well as post-hoc evaluation.

Despite these limitations, we have shown that % the use of a parameterized proposal network makes 
our method is highly flexible; it works with discrete design spaces, non-differentiable and implicit likelihood models. % Moreover, RL-sCEE does not require explicit likelihood information and is thus applicable when only implicit likelihood models are available. 
Surprisingly, our method even outperforms baselines that rely on explicit likelihoods.

Finally, our experiments show that the neural posterior proposal, $q_\kappa(\cdot)$, can learn significant structure that we know theoretically should be present in the true Bayesian posterior, such as symmetries or constraints on the support. This structure is learned entirely from data and self-guided experiments, with no inductive bias of any kind. And when the neural posterior proposal class may perform sub-optimally or additional information about the structure of the posterior over $\theta$ is known, $q_\kappa(\cdot)$ can be trivially swapped out for more efficiently parameterized proposal distributions.

\section{Impact Statement}

This paper presents work whose goal is to advance the field of Machine Learning. There are many potential societal consequences of our work, none which we feel must be specifically highlighted here.

\bibliography{main}

\newpage
\appendix
\onecolumn
\setcounter{theorem}{0}

\section{Proofs}

This appendix enumerates the proofs for the theorems, corollaries and other claims made in the main paper.

\subsection{Proof of~\cref{thm:bound}}\label{app:bound_proof}

Here we prove the main theorem of the paper, which is restated for convenience

\begin{theorem}
    Let $p(y \g \theta, d)$ be a probabilistic model with prior $p(\theta)$. For an arbitrary fixed design policy $\pi$ and sequence length $T$, the \gls{EIG} of using $\pi$ to design $T$ experiments is denoted $\EIG(\pi, T)$.
    Let $q(\theta \g h_T, \pi)$ be a proposal distribution over parameters $\theta$ conditioned on experimental history $h_T$, and the sCEE bound is
    \begin{equation}
        \sCEE(\pi, T) \coloneqq \E_{p(\theta, h_T \g \pi)} \left[ \log q(\theta \g h_T, \pi) \right] + \entropy\!\left[ p(\theta) \right] \label{eq:scee}
    \end{equation}
    we have that
    \begin{equation}
        \sCEE(\pi, T) \leq \EIG(\pi, T)
    \end{equation}
\end{theorem}

\begin{proof}
    From Theorem 1 of~\citet{foster2021deep} we have that the \gls{EIG} is:
    \begin{equation}
        \EIG(\pi, T) = \E_{p(h_T, \theta \g \pi)} \left[ \log p(h_T \g \theta, \pi) - \log p(h_T \g \pi) \right]
    \end{equation}
    This can be rewritten into a more convenient form:
    \begin{align}
        \EIG(\pi, T) &= \E_{p(h_T, \theta \g \pi)} \left[ \log\frac{ p(h_T \g \theta, \pi)}{ p(h_T \g \pi)} \right] = \E_{p(h_T, \theta \g \pi)} \left[ \log\frac{ p(h_T, \theta \g \pi)}{ p(h_T \g \pi) p(\theta)} \right] \\
        &= \E_{p(h_T, \theta \g \pi)} \left[ \log\frac{ p(\theta \g h_T, \pi) \cancel{p(h_T \g \pi)} }{ p(\theta) \cancel{p(h_T \g \pi)} } \right] = \E_{p(h_T, \theta \g \pi)} \left[ \log\frac{ p(\theta \g h_T, \pi) }{ p(\theta) } \right] \\
        &= \E_{p(h_T, \theta \g \pi)} \left[ \log p(\theta \g h_T, \pi) - \log p(\theta)  \right] \\
        &= \E_{p(h_T, \theta \g \pi)} \left[ \log p(\theta \g h_T, \pi) \right] + \entropy\!\left[ p(\theta) \right].\label{eq:eig}
    \end{align}
    We proceed to show that sCEE lower bounds this form. Consider the KL divergence between 2 conditional distributions given a fixed value $y$:
    \begin{equation}
        \KL{p(x \g y)}{q(x \g y)} = \E_{p(x \g y)} \left[ \log \frac{p(x \g y)}{q(x \g y)} \right]
    \end{equation}
    If $y$ is not fixed but random we then take an expectation:
    \begin{align}
        \E_{p(y)}\left[ \KL{p(x \g y)}{q(x \g y)} \right] &= \E_{p(x \g y)p(y)} \left[ \log \frac{p(x \g y)}{q(x \g y)} \right] \\
            &= \E_{p(x, y)} \left[ \log \frac{p(x \g y)}{q(x \g y)} \right] \\
            &= \E_{p(x, y)} \left[ \log p(x \g y) - \log q(x \g y) \right]
    \end{align}
    rearranging the sides gives
    \begin{align}
            \E_{p(x, y)} \left[ \log q(x \g y) \right] &= \E_{p(x, y)} \left[ \log p(x \g y) \right] - \E_{p(y)}\left[ \KL{p(x \g y)}{q(x \g y)} \right] \label{eq:difference}\\
            &\leq \E_{p(x, y)} \left[ \log p(x \g y) \right]
    \end{align}
    where the last line exploits the fact that the KL divergence is always non-negative. Plugging in $x = \theta; y = (h_T; \pi)$ yields the lower bound:
    
    \begin{equation}
        \E_{p(\theta, h_T, \pi)} \left[ \log q(\theta \g h_T, \pi) \right] \leq \E_{p(\theta, h_T, \pi)} \left[ \log p(\theta \g h_T, \pi) \right]
    \end{equation}
    For a known policy $\pi$ this becomes:
    \begin{equation}
        \E_{p(\theta, h_T \g \pi)} \left[ \log q(\theta \g h_T, \pi) \right] \leq \E_{p(\theta, h_T \g \pi)} \left[ \log p(\theta \g h_T, \pi) \right]
    \end{equation}
    Adding the prior entropy to both sides yields:
    \begin{equation}
        \E_{p(\theta, h_T \g \pi)} \left[ \log q(\theta \g h_T, \pi) \right] + \entropy\!\left[ p(\theta) \right] \leq \E_{p(\theta, h_T \g \pi)} \left[ \log p(\theta \g h_T, \pi) \right] + \entropy\!\left[ p(\theta) \right].
    \end{equation}
    Finally, plugging in \cref{eq:scee,eq:eig} completes the proof:
    \begin{equation}
        \sCEE(\pi, T) \leq \EIG(\pi, T)
    \end{equation}
\end{proof}

\subsection{Proof of corollaries}\label{app:corollaries}

In the main paper we state a corollary of the above theorem:
\begin{corollary}
    The bound is tight if and only if $p(\theta \g h_T, \pi) = q(\theta \g h_T, \pi)$, and the bias of the sCEE estimator is $-\E_{h_T} \left[ \KL{p(\theta \g h_T, \pi)}{q(\theta \g h_T, \pi)} \right]$
\end{corollary}

If we subtract the lower bound from the \gls{EIG} we get the difference:
\\
\begin{equation}
    \E_{p(\theta, h_T \g \pi)} \left[ \log p(\theta \g h_T, \pi) \right] - \E_{p(\theta, h_T \g \pi)} \left[ \log q(\theta \g h_T, \pi) \right].
\end{equation}
\\
From~\cref{eq:difference} it follows that this difference is
\begin{equation}
    -\E_{h_T} \left[ \KL{p(\theta \g h_T, \pi)}{q(\theta \g h_T, \pi)} \right]
\end{equation}
Since the KL divergence is always non-negative, this difference is $0$ and the bound is tight if and only if $\KL{p(\theta \g h_T, \pi)}{q(\theta \g h_T, \pi)} = 0$ for all realisations of $h_T$. This establishes both corollaries.

\subsection{Proof of convergence}\label{app:convergence_proof}

In the main paper we make the claim that a Monte Carlo estimator of the sCEE converges at a rate of $O(\frac{1}{\sqrt{n}})$, where $n$ is the number of MC samples. Since the prior is known, we can rely on standard MC convergence proofs for the prior entropy component. Thus we need only worry about a convergence proof for estimating the cross-entropy component $\E_{p(\theta, h_T \g \pi)} \left[ \log q(\theta \g h_T, \pi) \right]$.
We denote the cross-entropy as $H\left[ p(\theta \g h_T, \pi), q(\theta \g h_T, \pi) \right]$ and the MC estimator as
\begin{equation}
    \hat{\entropy}\!\left[ p(\theta \g h_T, \pi), q(\theta \g h_T, \pi) \right] = \frac{1}{n} \sum_{i=1}^n -\log q(\theta^i \g h_T^i, \pi)
\end{equation}
According to Theorem 5.1 of~\citet{mcallester2020formal}, if there is a minimum log-likelihood $F_{max}$ such that $\log q(\theta \g h_T, \pi) \geq F_{max}$,  then with probability at least $1-\delta$ we have that
\begin{equation}
    \left|\entropy\!\left[ p(\theta \g h_T, \pi), q(\theta \g h_T, \pi) \right] - \hat{\entropy}\!\left[ p(\theta \g h_T, \pi), q(\theta \g h_T, \pi) \right]\right| \leq F_{max}\sqrt{\frac{\log(\frac{2}{\delta})}{2n}}
\end{equation}
Thus the MC estimator converges to the true sCEE with high probability at the desired rate of $O(\frac{1}{\sqrt{n}})$.

\section{Relationship between sCEE and sACE}\label{app:sace}

\citet{foster2021deep} propose in the appendices a lower bound EIG estimator that relies on a parameterised proposal distribution that approximates the posterior $p(\theta \g h_T)$. They called this the \textit{sequential Adaptive Contrastive Estimation} (sACE):

\begin{equation}
    \E_{p(\theta_0, h_T \g \pi)q(\theta_{1:L}; h_T)} \left[ \log \frac{p(h_T \g \theta_0, \pi)}{\frac{1}{L+1} \sum_{l=0}^L \frac{p(h_T \g \theta_l, \pi) p(\theta_l)}{q(\theta_l;h_T)}} \right]
\end{equation}

This is a contrastive bound where the contrastive samples are distributed according to the proposal distribution $\theta_{1:l} \sim q(\theta \g h_T)$. The construction and proof assume a minimum of $1$ contrastive sample. However, if we set $L=0$ in this expression, the sampling of contrastive samples from $q(\theta \g h_T)$ disappears and we get:

\begin{align}
    \E_{p(\theta_0, h_T \g \pi)} \left[ \log \frac{p(h_T \g \theta_0, \pi)}{\frac{1}{0+1} \sum_{l=0}^0 \frac{p(h_T \g \theta_l, \pi) p(\theta_l)}{q(\theta_l;h_T)}} \right] &=
    \E_{p(\theta_0, h_T \g \pi)} \left[ \log \frac{p(h_T \g \theta_0, \pi)}{\frac{1}{0+1} \frac{p(h_T \g \theta_0, \pi) p(\theta_0)}{q(\theta_0;h_T)}} \right] \\
    &= \E_{p(\theta_0, h_T \g \pi)} \left[ \log \frac{\cancel{p(h_T \g \theta_0, \pi)}}{\cancel{p(h_T \g \theta_0, \pi)}} \frac{q(\theta_0;h_T)}{p(\theta_0)} \right] \\
    &= \E_{p(\theta_0, h_T \g \pi)} \left[ \log \frac{q(\theta_0;h_T)}{p(\theta_0)} \right],
\end{align}
which is equivalent to the sCEE. Note that by avoiding the need for contrastive samples, the sCEE gains a considerable computational advantage. In the RL setting, the rewards depend on $q(\theta \g h_T)$ and hence need to be recomputed every time $q$ is updated. With the sACE estimator, this recomputation requires resampling the contrastive samples, increasing the computational effort by a factor of $O(L)$. Indeed, depending on memory constraints, it may not be possible to recompute an entire batch of rewards in a single vectorised operation. With the sCEE, however, reward recomputation requires only a single neural network pass.

In addition to the computational benefits, sCEE has the further advantage that it is compatible with implicit likelihood models, wherease sACE requires explicit models, since it includes the term $p(h_T \g \theta_0, \pi)$ in the numerator.

\section{Reinforcement learning algorithm}
\label{app:rl}

\citet{blau2022optimizing} have shown that the problem of learning an experiment design policy can be formulated as a special case of a \gls{MDP} called the SED-MDP. We follow their formulation for the reinforcement learning
algorithm in this paper, with the main difference being the use of the \gls{SCEE} reward and consequently the use of an approximate proposal
$q_\kappa(\theta \g h_T)$ parameterised as a conditional normalising flows 
neural network~\citep{winkler2019learning} with parameters $\kappa$.

The SED-MDP is a hidden parameter \gls{MDP} \citep{doshi2016hidden} which is a tuple of the form $\mathcal{M} = (\cS, \cA, \Theta, \mathrm{Tr}, \mathrm{R}, \gamma, P_\Theta)$ where,
\begin{itemize}
    \item $\cS$, the system state, corresponds to the space of histories, $\mathcal{H}$,
    \item $\cA$, the action space, and corresponds to the design space, $\mathcal{D}$,
    \item $\Theta$ is the space of unobserved parameters, $\theta$,
    \item $\mathrm{Tr} : \cS \times \cA \to \mathcal{P}_{\cS}(\cS)$ are the transition dynamics, and correspond to the pooled history encoder, $B_{\psi, t}$,
    \item $\mathrm{R} : \cS \times \cA \times \cS \to \mathbb{R}$ : is reward function, and corresponds to~\cref{eq:reward},
    \item $\gamma \in (0, 1]$ is a discount factor applied to rewards,
    \item $P_\theta$ is a prior over the parameter space, and is chosen to be $p(\theta)$ at the beginning of the episode.
\end{itemize}
The aim is to then find a policy, $\pi^* : \cS \to \mathcal{P}_{\cA}(\cA)$, that maximizes the expected discounted return,
\begin{align}
    J(\pi) \coloneqq \E\left[\sum_{t=1}^T \gamma^{t-1} \mathrm{R}(s_{t-1}, a_{t-1}, s_t) \right],
\end{align}
over a time horizon, $T$, and the expectation is over all probabilistic quantities in the tuple.

% Here $\Theta$ is the space of unobserved parameters that are fixed over the course of a rollout, and correspond to the parameters $\theta$ of the likelihood model in design of experiments, where $P_\Theta$ is the prior. The reward function and transition dynamics are parameterised by $\theta$. The key idea is to map experimental designs to policy actions $a_{t-1} = d_t$ and history information to system states:
% \\
% \begin{align}
%     \begin{split}
%         s_t &= (B_{\psi,t}, C_t, y_t), \\
%         y_t &\sim p(y_t \g d_t, \theta_0) , \\
%         B_{\psi,t} &= B_{\psi,t-1} + ENC_\psi(d_t, y_t) \text{ and}\\
%         C_t &= C_{t-1} \odot \left[ p(y_t \g \theta_l, d_t) \right]_{l=0}^L,
%     \end{split}
% \end{align}
% \\
% where $ENC_\psi(d_t, y_t)$ is a learned encoding of the most recent experiment, $\psi$ are the parameters of the encoding and $\odot$ is the Hadamard product. If the discount is set to $\gamma = 1$ and the reward function is set to $R(s_{t-1}, a_{t-1}, s_t, \theta) = \log p(y_t \g \theta_0, d_t) - \log (C_t \cdot \mathbf{1}) + \log (C_{t-1} \cdot \mathbf{1})$, then the expected discounted return is identical to the \gls{SPCE} and the design policy can be optimised with any \gls{RL} algorithm we choose.
%
% Policy and critic networks $\pi_\phi$ and $\mathcal{C}_\chi$ can be updated following the rules of the RL algorithm of our choice, using mini-batches of either off-policy or on-policy samples.
The posterior network $q_\kappa(\cdot)$ can be updated by using the same mini-batches 
to maximise the log-likelihood of the observations under our posterior model. 
Note that this means rewards are now no longer fixed but depend on $q_\kappa(\cdot)$, and change with every update of $\kappa$. The computational cost thus incurred can be minimised by lazy evaluation~\citep{bloss1988code}: we only update each reward when we are about to use it to update the policy and critic networks of the RL agent. The procedure is summarised in~\cref{alg:scee-rl}, and we give more details about this procedure in the following sections.

% \begin{algorithm}[t]
% 	\caption{RL-sCEE}
% 	\begin{algorithmic}
%         \label{alg:scee-rl}
%         \STATE {\bfseries Input:} $\mathcal{M}$: SED-MDP, $L_\pi$: policy loss function, $L_\mathcal{C}$: critic loss function
%         \STATE Initialise replay buffer $\cB$
%         \WHILE{convergence criterion not reached}
%             \STATE Generate rollouts $(s_{0:T}, a_{0:T}, \theta)^{1:N}$ using $\cM$ and $\pi$ and push to $\cB$.
%             \STATE Sample mini-batch of size $mb$ from $\cB$
%             \STATE Compute posterior loss $L_q = -\frac{1}{mb}\sum_{i=1}^{mb}\log q_\kappa(\theta^i \g B_{\psi,t}^i, \pi_\phi)$
%             \STATE Take gradient step to minimise $\nabla_\kappa L_q$
%             \STATE Compute rewards for mini-batch using~\cref{eq:reward}
%             \STATE Use mini-batch and rewards to compute $L_\pi$ and $L_\mathcal{C}$
%             \STATE Take gradient step to minimise $\nabla_\phi L_\pi$ and $\nabla_\chi L_\mathcal{C}$
%         \ENDWHILE
%     \end{algorithmic}
% \end{algorithm}

\subsection{Simultaneous policy and reward learning}
\label{sec:practical-considerations}

We propose to learn the design policy network $\pi_\phi$ and the proposal distribution $q_\kappa(\cdot)$ from data simultaneously. Since the reward function depends on $q_\kappa(\cdot)$, and the objective function of $q_\kappa(\cdot)$ in turn depends on $\pi_\phi$, this leads to inherent instability, similar to the ``deadly triad'' that is often observed in value-based reinforcement learning~\citep{van2018deep}. We therefore apply several stabilisation mechanisms to prevent the neural network estimators from diverging.

\textbf{Target posterior network:} similar to the use of target Q-networks as introduced by~\citet{lillicrap2016continuous}, we maintain a primary posterior network $q_\kappa$ and a target network $q_\kappa^\prime$. The primary network $q_\kappa$ is updated using gradient descent in every iteration of the algorithm, but is not used directly to compute rewards. Instead, the target network $q_\kappa^\prime$ is used to compute~\cref{eq:reward}, and its weights are periodically updated to maintain a moving average:
\begin{equation}
    \kappa^\prime \gets \kappa^\prime \cdot (1 - \tau) + \kappa \cdot \tau
\end{equation}
where $\tau \in (0,1)$ is a constant controlling the rate of change.

\textbf{Fixed % prior likelihood
initial posterior:} the reward definition of~\cref{eq:reward} assigns each experiment its own (estimated) information gain. 
The return of an entire trajectory is a telescoping sum that reduces to $\log q(\theta \g B_{\psi,T}, \pi_\phi) - \log q(\theta \g B_{\psi, 0})$, and the expected return over infinitely many trajectories recovers the \gls{SCEE}. Therefore, the component $q(\theta \g B_{\psi,0})$ of the first reward $r_0$ is the only contributor to the prior entropy term $\entropy[p(\theta)]$ of the sCEE. Since this term is constant w.r.t. all networks, we can simply ignore it when training as it does not change the optimal policy. Furthermore, learning the correct estimator for $q(\theta \g B_{\psi,0})$ that maps the null inputs to the prior $p(\theta)$ can be challenging. Therefore instead of learning this mapping for the empty first state, we assigned it a fixed value of $\log q(\theta \g B_{\psi,0}) \coloneqq 0$.

\subsection{Implementation}

We use the \gls{REDQ} soft actor critic method from~\cite{chen2020randomized}, specifically that given in Algorithm 1 in their paper. Accordingly the actor and critic losses, $L_\pi$ and $L_\mathcal{C}$ respectively, are given on lines 10 and 12 in their Algorithm 1, using the reward we have defined in \cref{eq:reward}.

\section{Normalising flows on the probability simplex}

If we have a random variable with support on the canonical (open) simplex $\Delta_{k-1}$ rather than in $\cR^k$, additional caution is required for fitting a normalising flow to this RV. Since the $k^{th}$ dimension of the RV is fully determined by the first $k-1$ dimensions, the NF is free to fit this dimension with extremely high confidence, leading to an overestimation of log-likelihood of the entire RV.

The fix to this issue is rather involved. First, we exclude the $k^{th}$ dimension as input to the NF. Then, at the penultimate layer of a normalising flow, it implements the diffeomorphism  $\cF : \cR^{k-1} \rightarrow \cR^{k-1}$ i.e. the base distribution is a standard Gaussian and the resulting distribution can have support in the entire real space. Now we add a series of bijections that will produce a map $\cG : \cR^{k-1} \rightarrow \Delta_{k-1}$. Note that it is not enough simply to concatenate $1 - \sum_{i=1}^{k-1} \cF(x)_i$ with the intermediate vector $\cF(x)$ because we are not guaranteed that $0 \leq \cF(x)_k \leq 1 \  \forall k$ and that $\sum_{i=1}^{k-1} \cF(x)_i \leq 1$. First we must transform the output to ensure these properties:

\begin{align}
    &u = \cF(x)\\
    &v_i = \sigma(u_i) \label{eq:open-box}\\
    &w_i = v_i\left(1 - \sum_{j=1}^{i-1}w_j \right) \ \forall i\in[1, k-1] \label{eq:low-simplex}\\
    &\theta_i = \frac{w_i}{1 - \epsilon}. \label{eq:epsilon}
\end{align}

\Cref{eq:open-box} projects $\cR^{k-1}$ to the semi-open box $[0,1)^{k-1}$. \Cref{eq:low-simplex} projects this box to the $k-1$ dimensional simplex $\mathbf{s} = \{x : \sum_{i=1}^{k-1} x_i < 1 \ \text{and} \ 0 \leq x_k < 1 \ \forall i \in [1,k-1]\}$. This non-canonical simplex is in fact the equivalent of projecting the $k$-dimensional canonical simplex $\Delta_{k-1}$ down to $k-1$ dimensions. The simplex $\mathbf{s}$ can be lifted to $\Delta_{k-1}$ by assigning $w_k = 1 - \sum_{j=1}^{k-1}w_j$. However, we won't include this in the mapping $\cG$ because it makes the Jacobian low-rank and hence the inverse ill-defined. To avoid floating-point errors, each element of the RV actually has to be in the range $[\epsilon, 1-\epsilon]$ where $\epsilon$ is the machine epsilon.~\Cref{eq:epsilon} maps between this space and the actual canonical simplex.

The corresponding log-det-Jacobians are:

\begin{align}
    &\sum_{i=1}^{k-1} \log (0.99 \cdot v_i(1-v_i)), \\
    &\sum_{i=1}^{k-1} \log (1 - \sum_{j=1}^{i-1}w_j), \\
    &(1 - k) \cdot \log(1 - \epsilon).
\end{align}

The inverse $\cG^{-1}$ can be written compactly as:

\begin{equation}
    u_i = \sigma^{-1} \left( \frac{(1 - \epsilon) \cdot \theta_i}{1 - \sum_{j=1}^{i-1} (1 - \epsilon) \cdot \theta_j} \right) \quad \forall i \in [1,k-1].
\end{equation}

\section{Experiment details}
\label{app:exp}
This appendix describes the probabilistic models, hyperparameters, and all other details relating to the experiment design problems appearing in the paper.

We implemented our algorithm using Pyro \citep{bingham2018pyro} and normflows~\cite{normflows} along with the Garage framework \citep{garage} and the REDQ algorithm \citep{chen2020randomized} for reinforcement learning. For complete details about algorithms and hyperparameters, see~\cref{app:algos}.
%To evaluate the \gls{EIG} we used contrastive estimators with $L=1\textrm{E}8$, a number of contrastive samples that is impractical for learning, but achieves better estimation than sCEE in most problems we investigated.

\subsection{Synthetic data -- EIG for conjugate priors}\label{app:conjugate}

For an isotropic Gaussian prior $\cN(\mu_0, \sigma_0\mathbf{I_k})$ and Gaussian likelihood with known isotropic covariance $\sigma\mathbf{I_k}$, the posterior after $n$ observations is an isotropic Gaussian with covariance:
\begin{align}
    \Sigma_{post} &= (\sigma_0^{-1}\mathbf{I}_k + n\sigma^{-1}\mathbf{I}_k)^{-1} \\
                  &= (\sigma_0^{-1} + n\sigma^{-1})^{-1} \mathbf{I}_k
\end{align}
The mean of the posterior is unimportant to us as it does not affect the entropy:
\begin{align}
    \entropy_{post} &= \frac{k}{2} + \frac{k}{2}\log(2\pi) + \frac{1}{2}\log (|\Sigma_{post}|) \\
             &= \frac{k}{2} + \frac{k}{2}\log(2\pi) - \frac{k}{2}\log(\sigma_0^{-1} + n\sigma^{-1}).
\end{align}
Therefore the entropy is independent of the designs and we can compute the entropy of the ``optimal'' policy by subtracting the posterior entropy from the prior entropy:
\begin{align}
    I_n(\pi) &= \entropy[\cN(\mu_0, \sigma_0\mathbf{I_k})] - \entropy_{post} \\
             &= \cancel{\frac{k}{2}} + \cancel{\frac{k}{2}\log(2\pi)} + \frac{k}{2}\log (\sigma_0) - \cancel{\frac{k}{2}} - \cancel{\frac{k}{2}\log(2\pi)} + \frac{k}{2}\log(\sigma_0^{-1} + n\sigma^{-1}) \\
             &= \frac{k}{2} (\log (\sigma_0) + \log(\sigma_0^{-1} + n\sigma^{-1})) \\
             &= \frac{k}{2} \log (1 + n \frac{\sigma_0}{\sigma})
\end{align}

Thus we can create an \gls{EIG} estimation problem with an \gls{EIG} of our choice by setting $k, n, \sigma_0$ and $\sigma$ appropriately. In our experiments sCEE was trained for $10,000$ epochs, and was exposed to $10,000$ data points in each epoch. Each estimator was evaluated using $1,000$ Monte Carlo samples.

\subsection{Constant elasticity of substitution}\label{app:ces}

We evaluate a design problem in behavioural economics where we must estimate the parameters of a Constant Elasticity of Substitution (CES) utility function~\cite{baltas2001utility}. In this experiment economic agents compare $2$ baskets of goods and give a rating on a sliding scale from $0$ to $1$. Each basket consists of $k$ different goods with different value. We set $k=3$.

The outcome is the relative preference of a test subject in the range $[0,1]$, as determined by the agent's CES utility function, and the specific values of its parameters $\theta = \{\rho, \alpha, u\}$, with $\rho \in [0,1]$, $ \alpha \in \Delta_3$ and  $u > 0$.

The designs are vectors $d = (x, x^\prime)$ where $x, x^\prime \in [0, 100]^k$ are the baskets of goods. The latent parameters of the likelihood and their priors are:

\begin{align}
    \rho &\sim Beta(1,1)\\
    \alpha &\sim Dirichlet(\mathbf{1}_k)\\
    \log u &\sim \cN(1,3).
\end{align}

The probabilistic model is:

\begin{align}
    U(x) &= \left(\sum_i x^\rho_i\alpha_i\right)^{1/\rho}\\
    \mu_\eta &= (U(x) - U(x^\prime)) u\\
    \sigma_\eta &= (1 + ||x - x^\prime||) \tau\cdot u\\
    \eta &\sim \cN(\mu_\eta, \sigma^2_\eta)\\
    y &= clip(sigmoid(\eta), \epsilon, 1-\epsilon),
\end{align}

In our experiments we used the following hyperparameters:
\begin{table}[htb]
\vskip 0.15in
\begin{center}
\begin{small}
\begin{sc}
\begin{tabular}{ll}
\toprule
Parameter & Value \\
\midrule
$k$    & $3$ \\
$\tau$    & $0.005$  \\
$\epsilon$    & $2^{-22}$  \\
\bottomrule
\end{tabular}
\end{sc}
\end{small}
\end{center}
\vskip -0.1in
\end{table}

\subsection{Prey population}\label{app:prey}

In this experiment an initial population of prey animals is left to survive for $\cT$ hours, and we measure the number of individuals consumed by predators at the end of the experiment. The designs are the initial populations $d = N_0 \in {1,2, \dots, 300}$. The latent parameters and priors are:
\begin{align}
    \log a &\sim \cN(-1.4, 1.35)\\
    \log T_h &\sim \cN(-1.4, 1.35),
\end{align}
where $a$ represents the attack rate and $T_h$ is the handling time.

The population changes over time according to a Holling's Type III model, which is a differential equation:
\begin{equation}
    \frac{dN}{d\tau} = - \frac{aN^2}{1 + aT_hN^2}. 
\end{equation}

And the population $N_\cT$ is thus the solution of an initial value problem. The probabilistic model is:
\begin{align}
    p_\cT &= \frac{d - N_\cT}{d}\\
    y &\sim Binom(d, p_\cT).
\end{align}

We used a simulation time of $\cT = 24$ hours.

\subsection{Source location}\label{app:source}

% In the next experiment we consider the problem of locating signal sources in space. Here we have $2$ sources with co-ordinates $s_1 = (x_1,y_1); s_2=(x_2,y_2)$, and designs consist of choosing the co-ordinates at which to take a noisy sample of signal magnitude. Signals decay with distance from the source and individual signals always superpose constructively. The full probabilistic model is described in~\cref{app:source}.

% \textbf{Implicit likelihood}: in both the CES and source location problem, we simulate an implicit likelihood by withholding the explicit likelihood values $p(y \g \theta, d)$ from the RL-sCEE and iDAD agents. One would expect that both agents would perform worse than the RL-sPCE and DAD agents, which do exploit explicit likelihood information. Indeed, the iDAD agent is the least performant of any method considered. However, our proposed RL-sCEE method outperforms all baselines, both in the source location and CES problems, in spite of not having access to explicit likelihoods.

In this experiment there are $n$ sources embedded in $k$-dimensional space that emit independent signals. the designs are the co-ordinates at which to measure signal intensity, and we restrict the space to $d \in [-4, 4]^k$. The total intensity at any given co-ordinate $d$ in the plane is given the sum of individual signals:
\begin{equation}
    \mu(\theta, d) = b + \sum_i\frac{1}{m + ||\theta_i - d||^2},
\end{equation}
where $b, m > 0$ are the background and maximum signals, respectively, $||\cdot||^2$ is the squared Euclidean norm, and $\theta_i$ are the co-ordinates of the $i^{th}$ signal source. The probabilistic model is:

\begin{equation}
    \theta_i \sim \cN(0, I); \quad \log y \g \theta, d \sim \cN(\log(\mu(\theta, d), \sigma),
\end{equation}

i.e. the prior is unit Gaussian and we observe the log of the total signal intensity with some Gaussian observation noise $\sigma$. The hyperparameters we used are:
\begin{table}[htb]
\vskip 0.15in
\begin{center}
\begin{small}
\begin{sc}
\begin{tabular}{lr}
\toprule
Parameter & Value \\
\midrule
$n$    & $2$ \\
$k$    & $2$ \\
$b$    & $1\textrm{E}-1$  \\
$m$    & $1\textrm{E}-4$  \\
$\sigma$    & $0.5$  \\
\bottomrule
\end{tabular}
\end{sc}
\end{small}
\end{center}
\vskip -0.1in
\end{table}

\section{Algorithm experimental details}\label{app:algos}

This appendix provides the implementation details for all design of experiment algorithms used in the paper.

\subsection{RL-sCEE}

We used the implementation of REDQ from~\citet{blau2022optimizing} as the basis of our algorithm, although we limited the ensemble size to $N=2$. Normalising flows were implemented using the normflows~\cite{normflows} package, which we extended to create a conditioned version of realNVP~\cite{dinhdensity}. In every experiment we used a normalising flow with $6$ layers, and the parameter map is a $2$-layer neural network with sizes $(128,128)$. Both normalising flows and policies  use a permutation invariant representation similar to~\citet{ivanova2021implicit}, including a single self-attention layer with $8$ attention heads.

Additional hyperparameters are listed in the table below, and are largerly derived from~\citet{blau2022optimizing}:

\begin{table}[htb]
\vskip 0.15in
\begin{center}
\begin{small}
\begin{sc}
\begin{tabular}{llll}
\toprule
Parameter & Source Location & CES & Prey Population \\
\midrule
training iterations    & $1\textrm{E}5$ & $1\textrm{E}5$  & $2\textrm{E}4$  \\
$T$    & $30$ & $10$ & $10$  \\
$\gamma$    & $0.9$ & $0.9$  & $0.95$  \\
$\tau$    & $1\textrm{E}-3$ & $5\textrm{E}-3$ & $1\textrm{E}-2$  \\
policy learning rate    & $1\textrm{E}-4$ & $3\textrm{E}-4$ & $1\textrm{E}-4$  \\
critic learning rate    & $3\textrm{E}-4$ & $3\textrm{E}-4$ & $1\textrm{E}-3$  \\
buffer size    & $1\textrm{E}7$ & $1\textrm{E}7$ & $1\textrm{E}6$  \\
\bottomrule
\end{tabular}
\end{sc}
\end{small}
\end{center}
\vskip -0.1in
\end{table}

For the Source Location experiments where $d=\{6, 10\}$, we had to reduce the buffer size from 1E7 to 1E6 in order to fit the estimators in the available 16GB of GPU memory.

\subsection{RL-sPCE}

We used the implementation of~\citet{blau2022optimizing}, which is available at \url{https://github.com/csiro-mlai/rl-boed}. We kept all hyperparameters and network architectures the same, with the exception of adding a self-attention layer to the policy network. This layer is identical to the one described in the previous section. We did not find that adding attention lead to significant change in performance, but included it in order to maintain a fair comparison with the RL-sCEE implementation.

In particular, we used $L=1\textrm{E}5$ contrastive samples for training. Not only is it the value used by~\citet{blau2022optimizing}, but is also pushing the limits of the number of samples that can be used in a reasonable amount of time. Since tens of millions of simulated experiments have to be run to train a single agent, we must leverage vectorisation over multiple sequences of experiments in parallel. Although in the evaluation we used $L=1\textrm{E}8$ samples, this only allows a single experiment at a time to fit in a GPU, and requires multiple seconds per experiment. It would require several years to train a single agent in this manner.

\subsection{DAD and iDAD}

For these baselines we used the implementations of the original papers, which are available at \url{https://github.com/ae-foster/dad} and \url{https://github.com/desi-ivanova/idad}, respectively. We kept the default hyperparameters of those implementations. The only exception is for iDAD on the source location problem, which we found unstable for a sequence of $T=30$ experiments. We therefore used early stopping, and stopped learning at $40k$ iterations instead of the original $100k$.

\subsection{SMC-ED}

We used the implementation made available in \url{https://github.com/csiro-mlai/rl-boed}, which in turn uses the R language implementation of~\cite{moffat2020sequential} and executes it from within a Python script by using the $rpy2$ bindings. The original R code is available at \url{https://github.com/haydenmoffat/sequential_design_for_predator_prey_experiments}.

\section{Hardware details}

SMC-ED experiments were run on a desktop machine with an Intel i7-10610U CPU and no GPU. All other experiments were run in a high-performance computing cluster, using a single node each with $4$ cores of an Intel Xeon E5-2690 CPU and an Nvidia Tesla P100 GPU with 16GB of VRAM.

\section{Ablation study of stabilisation mechanisms}

\begin{figure}[htb]
\vskip 0.1in
\begin{center}
\centerline{
    \includegraphics[width=0.5\columnwidth]{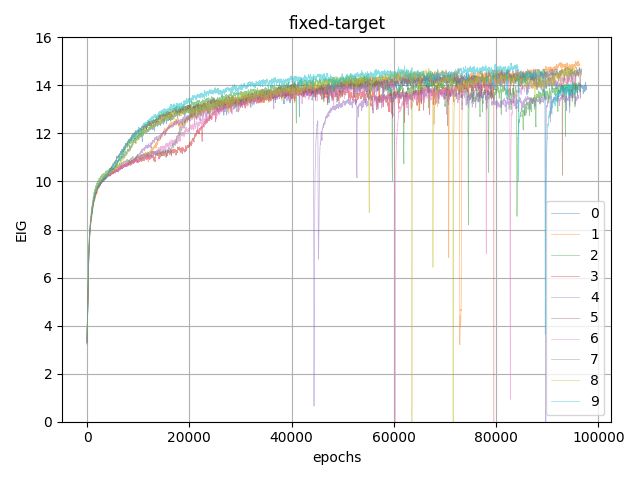}
    \includegraphics[width=0.5\columnwidth]{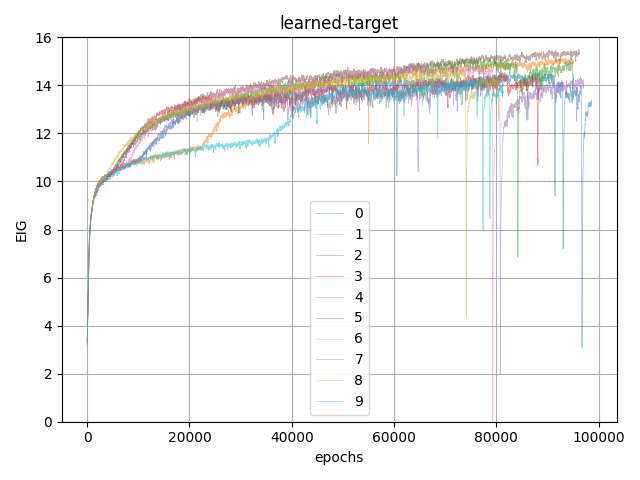}
}
\centerline{
    \includegraphics[width=0.5\columnwidth]{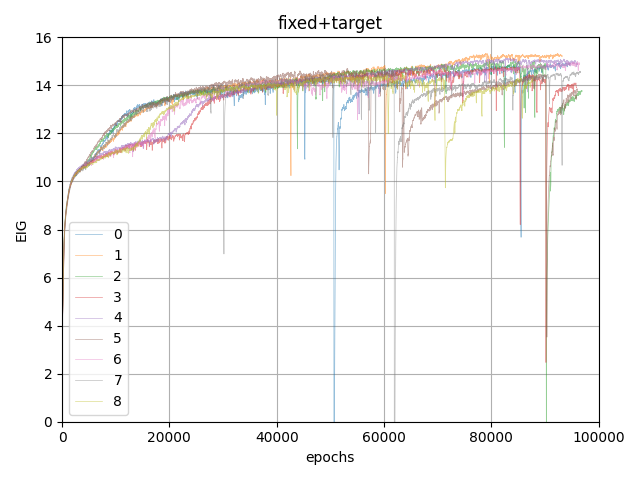}
    \includegraphics[width=0.5\columnwidth]{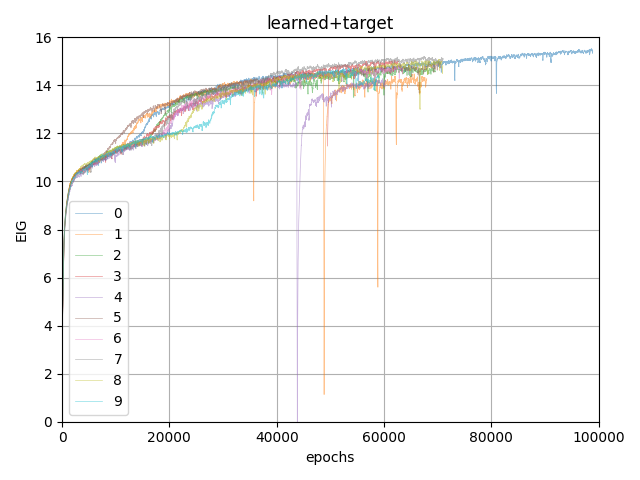}
}
\caption{Ablation studies for the stabilisation mechanisms.}
\label{fig:ablation}
\end{center}
\vskip -0.2in
\end{figure}

To evaluated the stabilisation mechanisms incorporated in the implementation of RL-sCEE, we conduct an ablation study where we remove either the target posterior network, the fixed initial posterior, or both.  The results can be seen in~\cref{fig:ablation}, with each variant replicated $10$ times, using common random seeds between different variants (e.g. the blue trendline labeled "0" represents the same random seed in all $4$ plots).

It is clear that the removal of the target network causes significant degradation in performance, with many replications converging to a lower final performance or even peaking early and then decreasing in EIG. On the other hand, the use of a fixed initial posterior doesn't seem to offer a clear advantage over a learned one.

\section{Additional results}\label{app:results}

\begin{figure}[htb]
\vskip 0.1in
\begin{center}
\centerline{
    \includegraphics[width=\columnwidth]{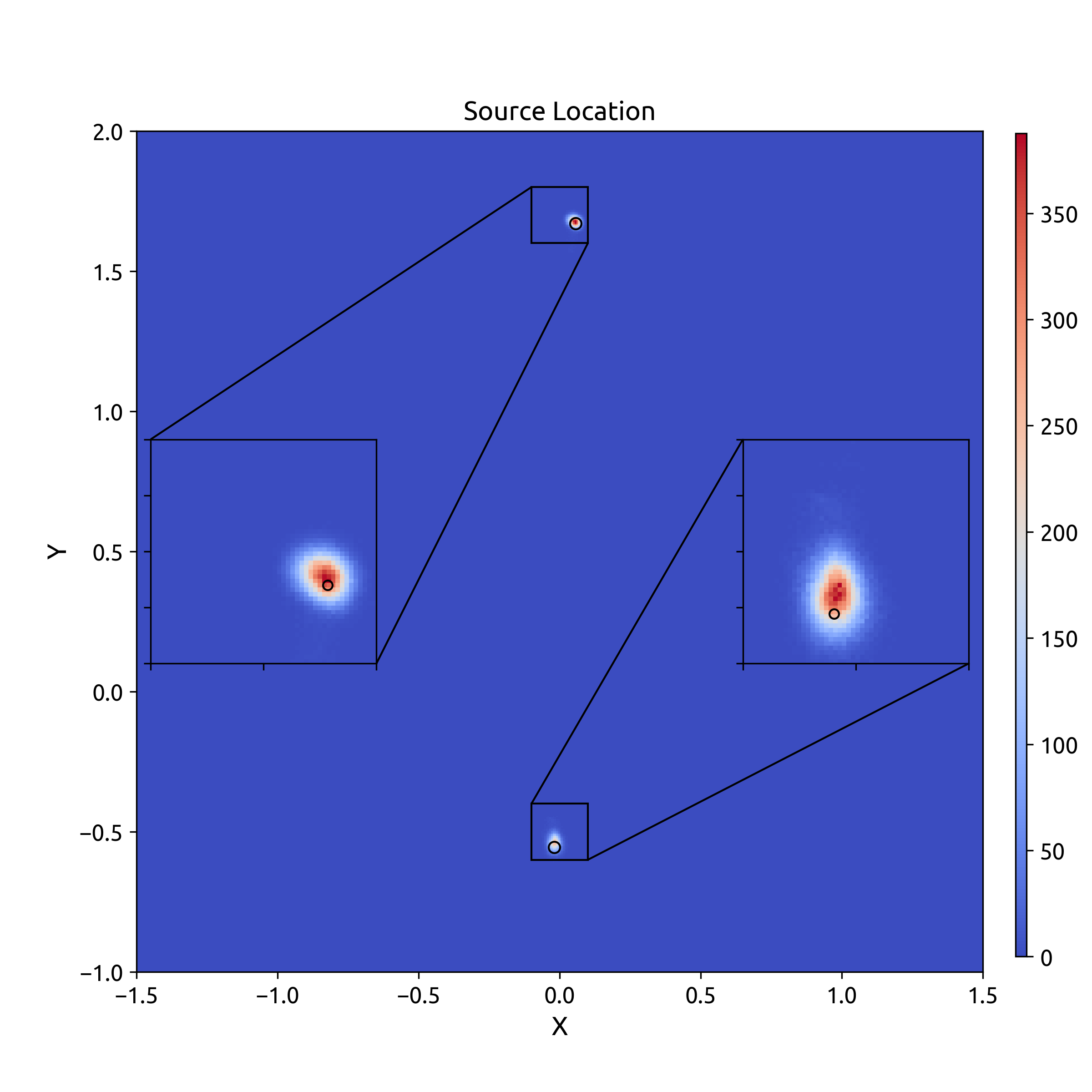}
}
\caption{Posterior for the source location  problem. Computed from $1\textrm{E}5$ samples. Black rings denote the true co-ordinates of signal sources.}
\label{fig:extra_post}
\end{center}
\vskip -0.2in
\end{figure}

\end{document}